\documentclass[runningheads]{llncs}

\usepackage{eccv}

\usepackage{eccvabbrv}

\usepackage{graphicx}
\usepackage{booktabs}
\usepackage{tabularx}

\usepackage[accsupp]{axessibility}  

\usepackage[pagebackref,breaklinks,colorlinks,citecolor=eccvblue]{hyperref}

\usepackage{orcidlink}

\usepackage{amssymb}
\usepackage{multirow}

\definecolor{codeblue}{rgb}{0.25,0.5,0.5}
\definecolor{nvgreen}{rgb}{0.92, 0.97, 0.85}

\definecolor{navyblue}{HTML}{0071BC}
\definecolor{hotpink}{HTML}{FF0080}

\definecolor{mygreen}{HTML}{FFD500}
\definecolor{myred}{rgb}{1, 0.9, 0.9}
\definecolor{mygray}{gray}{0.95}
\definecolor{mydarkblue}{rgb}{0,0.08,1}
\definecolor{mydarkred}{rgb}{0.8,0.02,0.02}
\definecolor{mydarkorange}{rgb}{0.40,0.2,0.02}
\definecolor{mypurple}{RGB}{111,0,255}
\definecolor{mygold}{rgb}{0.75,0.6,0.12}
\definecolor{mydarkgray}{rgb}{0.66, 0.66, 0.66}
\definecolor{mydarkgreen}{rgb}{0.02,0.6,0.02}
\definecolor{mygray}{gray}{0.9}
\definecolor{keynotegreen}{rgb}{0.04,0.52,0}
\definecolor{keynoteyellow}{rgb}{1,0.68,0}
\definecolor{LightCyan}{rgb}{0.88,1,1}
\definecolor{tabfirst}{rgb}{1, 0.7, 0.7}
\definecolor{tabsecond}{rgb}{1, 0.85, 0.7} 
\definecolor{tabthird}{rgb}{1, 1, 0.7} 
\definecolor{rbtred}{rgb}{255, 0, 0}

\usepackage[table]{xcolor}
\usepackage{multirow, colortbl, booktabs, graphicx}
\usepackage{subcaption}
\usepackage{xspace}
\usepackage{lipsum}
\usepackage{pifont}
\usepackage{listings}
\usepackage{wrapfig}
\usepackage{amssymb}
\usepackage{pdfrender}
\usepackage{enumitem}
\usepackage{adjustbox}

\definecolor{JSViolet}{RGB}{71,15,244}
\definecolor{JSRed}{RGB}{205,44,78}
\definecolor{RowHighlight}{gray}{0.9}
\definecolor{skyblue}{RGB}{212,239,251}

\newcommand{\xmark}{\textcolor{JSRed}{\ding{55}}}
\definecolor{baselinecolor}{HTML}{EEEEEE}

\newcommand{\rvx}{\mathbf{x}}

\newcommand{\rvz}{\mathbf{z}}
\newcommand{\rva}{\mathbf{a}}

\newcommand{\sname}{SpatialBoost\xspace}

\begin{document}

\title{SpatialBoost: Enhancing Visual Representation through Language-Guided Reasoning} 

\titlerunning{Enhancing Visual Representation through Language-Guided Reasoning}

\author{
Byungwoo Jeon\inst{1}\thanks{Equal contribution.}\orcidlink{0009-0006-9448-2582} \and
Dongyoung Kim\inst{1,2}\textsuperscript{$\star$}\orcidlink{0009-0001-4209-5314} \and
Huiwon Jang\inst{1,2}\orcidlink{0009-0005-1162-8631} \and \\
Insoo Kim\inst{1,3}\orcidlink{0000-0002-9928-778X} \and
Jinwoo Shin\inst{1,2}
}

\authorrunning{B. Jeon et al.}

\institute{
$^1$KAIST \quad $^2$RLWRLD \quad $^3$NAVER Cloud \\
\email{\{imbw2024,kingdy2002,jinwoos\}@kaist.ac.kr}
}

\maketitle

\begin{abstract}

Despite the remarkable success of large-scale pre-trained image representation models (i.e., vision encoders) across various vision tasks, they are predominantly trained on 2D image data and therefore often fail to capture 3D spatial relationships between objects and backgrounds in the real world, constraining their effectiveness in many downstream applications. To address this, we propose \sname, a scalable framework that enhances the spatial awareness of existing pre-trained vision encoders by injecting 3D spatial knowledge expressed in linguistic descriptions. 
The core idea involves converting dense 3D spatial information from 2D images into linguistic expressions, which is then used to inject such spatial knowledge into vision encoders through a Large Language Model (LLM). To this end, we adopt a multi-turn Chain-of-Thought (CoT) reasoning process that progressively incorporates dense spatial knowledge and builds hierarchical spatial understanding. To validate effectiveness, we adapt \sname to state-of-the-art vision encoders such as DINOv3, and evaluate its performance gains on a wide range of benchmarks requiring both 3D perception and general vision abilities. For instance, \sname improves DINOv3 performance from 55.9 to 59.7 mIoU on ADE20K, achieving state-of-the-art performance with 3.8\% gain over the pre-trained DINOv3. \href{https://rootyjeon.github.io/spatial-boost/}{Project page}.


\end{abstract}
\section{Introduction}
\label{sec:intro}

Pre-trained image representation models~\cite{he2020momentum, donahue2019large, chen2020generative, dosovitskiy2020image, li2023mage, assran2023self} have shown remarkable success in various downstream tasks, such as image classification~\cite{krizhevsky2009learning,cui2018large}, semantic segmentation~\cite{lin2014microsoft,zhou2019semantic}, monocular depth prediction~\cite{silberman2012indoor, geiger2012we}, and vision-language understanding~\cite{antol2015vqa, hudson2019gqa}. The core idea behind these successes is extracting transferrable representation from large-scale image datasets such as ImageNet~\cite{deng2009imagenet}, enabling the model to understand semantic information within images that is significantly useful for various downstream tasks. 

Despite their success, these models are predominantly trained on 2D images and hence face a fundamental challenge in acquiring 3D spatial awareness capabilities. Consequently, large vision language models struggle to discern 3D spatial relationships between objects in images~\cite{liu2023visual, fu2024blink, wang2025picture, cheng2025spatialrgpt}, and demonstrate sub-optimal performance in vision-based robotic control tasks compared to approaches that directly utilize 3D information~\cite{ze20243d, ke20243d, zhen20243d}. To address these limitations, na\"ive approaches are to train vision models on multi-view images that inherently encode spatial information~\cite{zhang2024monst3r, wang2024dust3r, charatan2024pixelsplat}. While these approaches have shown promise in robot control tasks~\cite{seo2023multi, sermanet2018time}, their broader applicability remains constrained by the need to use carefully curated data~\cite{yu2023mvimgnet} or obtain multi-view datasets from simulation environments~\cite{savva2019habitat}, creating significant limitations for scaling up these approaches. These challenges highlight the need for a novel framework that enables effective learning of 3D information with substantially less data.

\begin{figure*}[t]
  \centering
\centerline{\includegraphics[width=\textwidth]{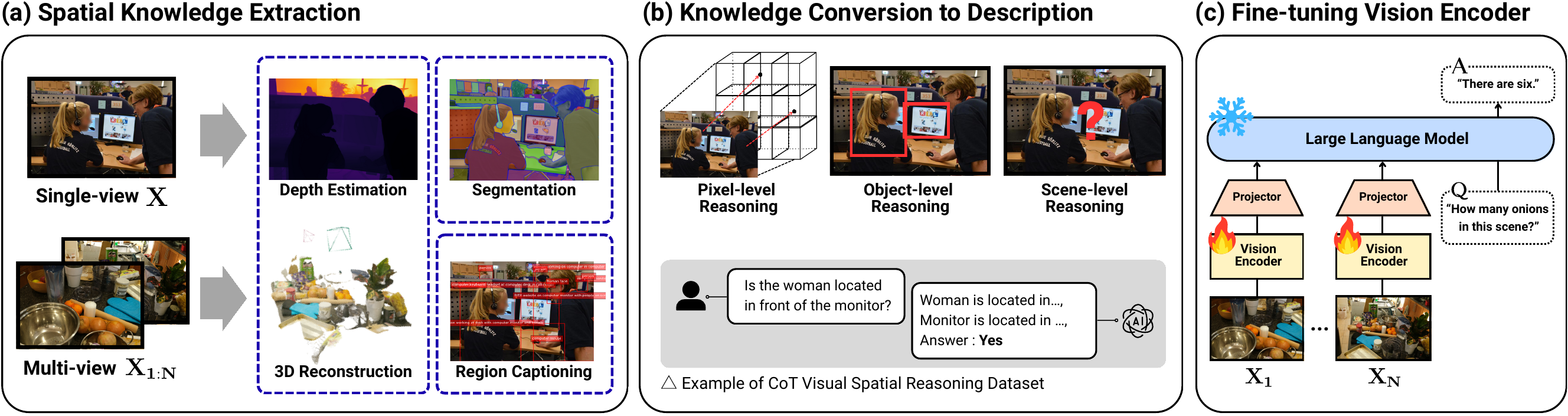}}
\vspace{-0.10in}
\caption{\textbf{Overview of \sname.} We enhance spatial and geometric understanding of pre-trained vision encoders by leveraging language-guided spatial reasoning. \sname consists of (a) spatial knowledge extraction through depth estimation, 3D reconstruction, segmentation, and region captioning, (b) converting spatial knowledge into multi-turn spatial reasoning from pixel to scene levels, and (c) building a spatial-aware vision encoder with LLM using generated data in (b).} 
\label{fig:overview}
\vspace{-0.4cm}
\end{figure*}

To address this problem, we note that vision models specialized for individual tasks are able to infer object positions and point depths from standard 2D images. These extracted cues make it possible to extend spatial information by modeling geometric relationships between objects in a scene.
We hypothesize that such spatial information can be systematically converted into explicit representations by leveraging language. 
Moreover, since language naturally composes information in a sequential and structured form, this property allows the construction of labels that capture dense spatial relationships within a scene.

Importantly, several recent works have shown that language can serve as a scalable supervision signal for learning visual representations~\cite{fini2025aimv2, jose2025dinomeetstext, bolya2025perception}. While these approaches primarily focus on semantic understanding or object localization, they highlight the potential of linguistic descriptions as scalable supervision for training vision encoders.

Based on these insights, we introduce \sname, a training framework that enhances the spatial understanding of pre-trained vision encoders by leveraging language-guided reasoning (see Figure~\ref{fig:overview}). 
We inject linguistically described spatial knowledge through decoder-based fine-tuning with Large Language Models (LLMs), enabling the framework to process large-scale textual descriptions in a single pass while handling both single-view and multi-view images.
In particular, to leverage this knowledge without forgetting the existing knowledge, we incorporate additional learnable parameters (\textit{i.e.}, dual-channel attention module) into the vision encoder and train only them while freezing the existing parameters. Furthermore, to incorporate dense spatial information in a structured manner, we present a multi-turn visual spatial reasoning approach that builds hierarchical spatial understanding through pixel-level, object-level, and scene-level sub-questions and answers.

To validate the effectiveness of our method, we apply \sname to state-of-the-art image encoders, including DINOv3~\cite{simeoni2025dinov3} and SigLIPv2~\cite{tschannen2025siglip2}, and evaluate them across a diverse set of vision tasks: monocular depth estimation, semantic segmentation, 3D scene understanding, vision-based robotic control, image classification, and image retrieval.
Our experiment first shows that \sname consistently improves performance on tasks requiring 3D spatial knowledge. For example, on the 3D scene understanding task, \sname improves DINOv3 by 3.5\%p (51.4\% $\rightarrow$ 54.9\%) on the SQA3D task from Lexicon3D Benchmark~\cite{man2024lexicon3d}. In addition, on depth estimation tasks, \sname improves SigLIPv2 from an RMSE score of 0.51 to 0.39 on NYUd linear probing. Moreover, we show that \sname even improves the performance of the vision encoders across all benchmarks, notably in image classification: \sname improves ImageNet linear probing performance of DINOv3 from 88.4\% to 90.2\%. In addition, DINOv3 with \sname achieves state-of-the-art performance across all evaluated tasks.
\section{Related Work}
\label{sec:relatedwork}

\subsection{Self-supervised Learning for Image Representation}

In earlier years, most approaches relied on supervised learning with large-scale labeled datasets to train models~\cite{deng2009imagenet, simonyan2014very, szegedy2014going, he2016deep}. However, the dependence on annotated data introduced scalability challenges due to label expense. To address this, self-supervised learning (SSL) has emerged as a dominant paradigm, leveraging unlabeled data to learn image representations. Contrastive learning methods, including  SimCLRv2~\cite{chen2020big}, MoCov3~\cite{chen2021empirical}, DINOv2~\cite{oquab2023dinov2}, and iBOT~\cite{zhou2021ibot}, are trained to distinguish between representations of augmented views of the same image and those of different images. Concurrently, mask prediction approaches such as BEiT~\cite{bao2021beit} and MAE~\cite{he2022masked}, learn representations by reconstructing masked portions of input images. While these methods excel at capturing rich semantic features within 2D images, they lack mechanisms to effectively encode 3D spatial knowledge. On the other hand, we overcome this limitation by enhancing image representations through a novel method that injects 3D spatial knowledge by utilizing language decoding.

\subsection{Multi-modal Learning for Image Representation}

The increasing prominence of multi-modal tasks has catalyzed the development of vision-language models that jointly represent visual and textual information. These models typically employ weakly supervised learning by leveraging text caption. Contrastive learning schemes, \eg, CLIP~\cite{radford2021learning}, SigLIP~\cite{zhai2023sigmoid} and OpenCLIP~\cite{cherti2023reproducible}, consist of vision and text encoders and are trained to align their representations in a shared embedding space.
Alternative methodologies like M3AE~\cite{geng2022multimodal}, jointly encode image patches and text tokens, employing masked prediction objectives to reconstruct both modalities. More recently, autoregressive formulations such as iGPT~\cite{chen2020generative}, have emerged, treating image patches and text tokens as sequential elements for predictive modeling. These approaches successfully enrich visial representations with semantic context derived from natural language descriptions. However, existing models necessitate joint pre-training of both modalities from scratch, imposing significant computational demands and preventing efficient adaptation of existing pre-trained models. Our method eliminates the need for joint text-image representation learning by using LLM, thereby enhancing pre-trained models with relevant linguistic information efficiently.

\subsection{Multi-View Learning for Image Representation}

Recent advances in vision tasks that require 3D spatial understanding and generation have increased the demand for effective 3D spatial representations~\cite{chen2024mvsplat, wu2024reconfusion, goyal2023rvt, shridhar2023perceiver}. Multi-view images from different camera viewpoints or video sequences serve as input for these tasks. Our focus is specifically on augmenting image representations with useful 3D information. Typically, following approaches similar to single-view image representation learning, multi-view data has been processed by converting images into patches for masked prediction such as MV-MWM~\cite{seo2023multi} or through contrastive learning methods~\cite{sermanet2018time}. Additionally, to learn 3D-related information more explicitly, approaches that predict 3D features from image representation~\cite{ke20243d,gervet2023act3d,ze20243d} have been proposed. These approaches have led to significant performance improvements in vision-based robot control. However, such methods are limited by multi-view data, making it difficult to develop them into pre-trained models for general 3D understanding. Our approach proposes a method to learn 3D spatial representations from both single-view and multi-view images, avoiding these limitations.
\section{Method}

In this section, we introduce \sname, a visual representation learning framework designed to improve vision encoders by injecting 3D spatial information expressed in natural language. We first present a multi-modal architecture that incorporates linguistically expressed visual information into the vision encoder through a dual-channel attention layer, ensuring that original visual features are preserved while 3D spatial information is fully exploited (see \Cref{sec:training_pipeline}). On top of this architecture, we design a Visual-Question-Answering (VQA) dataset that hierarchically disentangles 3D spatial relations from both single/multi-view images, enabling the vision encoder to learn spatial information more effectively (see \Cref{fig:overview} and \Cref{sec:method:CoT_data}).

\subsection{Training Pipeline}\label{sec:training_pipeline} 

\begin{figure*}[t]
  \centering
  \centerline{\includegraphics[width=\textwidth]{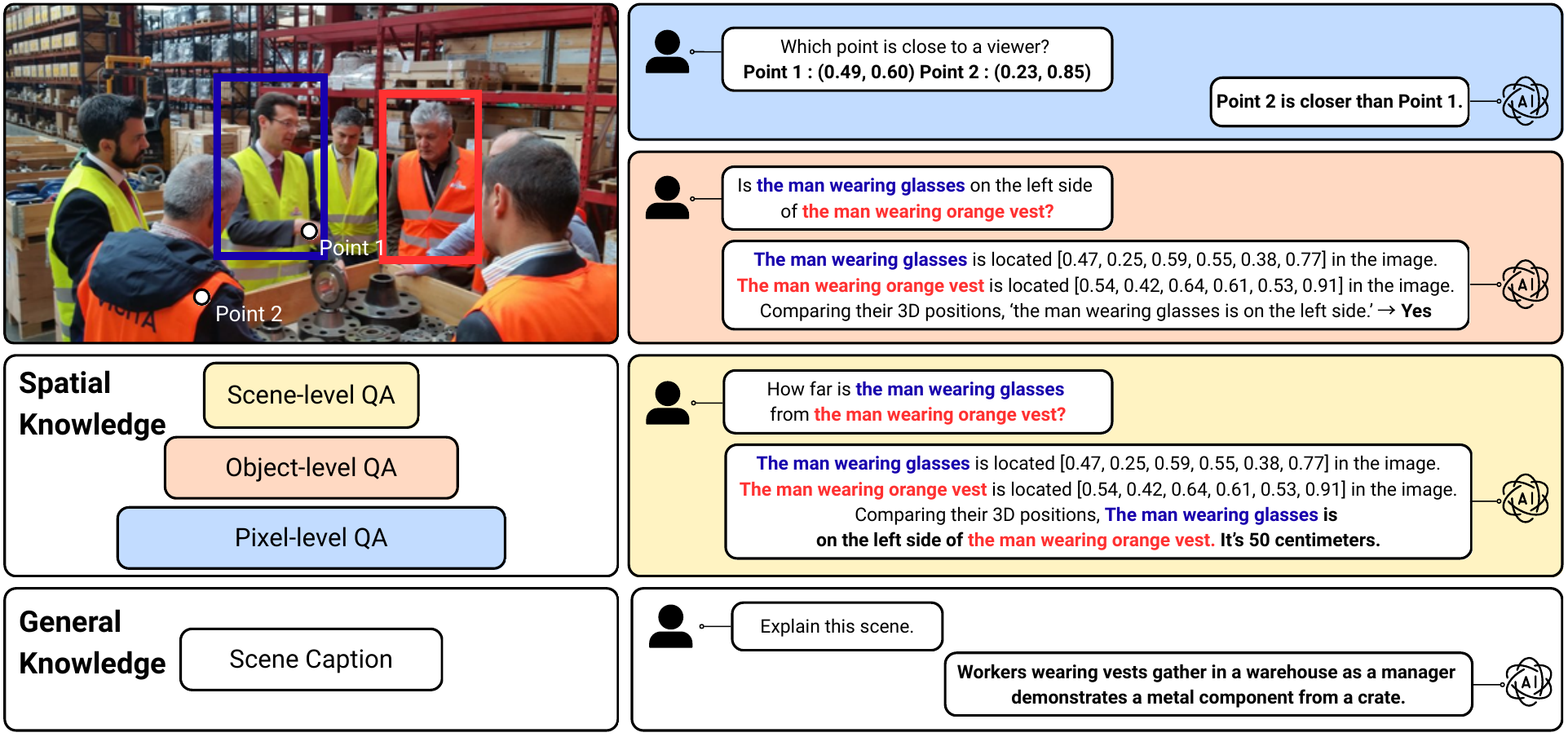}}
  \vspace{-0.1in}
   \caption{\textbf{Illustration of multi-turn visual spatial reasoning dataset}, exhibiting pixel-level, object-level, and scene-level reasoning QAs. 
   At the pixel-level, the QA task queries the 3D positions of points (\eg, via depth estimation). At the object-level, it extracts spatial properties of objects (\eg, by predicting bounding cubes or relative positions). At the scene-level, it determines the exact distances between multiple objects that require the rationales of the previous steps. At last, we add 2-turn for general scene caption. These are listed in order and constitute 12 multi-turn visual spatial reasoning conservation.}
   \label{fig:cot_data_gen2}
   \vspace{-0.4cm}
\end{figure*}

To train a vision encoder from rich spatial information encoded in large-scale linguistic expressions, our key idea is to utilize Large-Language Models (LLM) by constructing a multi-modal architecture composed of a vision encoder $f_V$, a trainable projection module $g_P$, and the LLM $f_L$.
However, without proper alignment between visual and textual representations, the training signals from the LLM cannot effectively propagate back to the vision encoder, making the learning process ineffective. To fully exploit language supervision, we begin by aligning the visual encoder with the textual embedding space of the LLM. Specifically, we adopt LLaVA~\cite{liu2023llava}, a two-stage training for the alignment: feature alignment (Stage 1) and visual instruction tuning (Stage 2). After the alignment, we introduce a training framework that uses a language-guided reasoning dataset to fine-tune the vision encoder (Stage 3).
Notably, direct full fine-tuning in this final stage would lead to catastrophic forgetting of the pre-trained knowledge embedded in the vision encoder. To address this challenge, we introduce \textit{dual-channel attention} layers that enable the model to acquire spatial understanding while preserving its original representational capabilities.

Formally, given an input image $\rvx$ and multi-turn conversation data $(\rvx_\mathtt{q}^1, \rvx_\mathtt{a}^1, $ \\ $\cdots, \rvx_\mathtt{q}^T, \rvx_\mathtt{a}^T)$ from question-answering (QA) pairs $(Q_\rvx, A_\rvx)$, we first encode $\rvx$ to obtain visual features $\rvz_\mathtt{v} = f_V(\rvx)$, which are mapped into the token embedding space via $g_{P}(\rvz_{\mathtt{v}})$. These visual tokens are then concatenated with text tokens and fed into the LLM.
Given the multi-turn conversation data and input image, we optimize the model through autoregressive loss. Our training pipeline consists of three stages and all stages are trained with supervised fine-tuning (SFT) loss. We describe each stage in the following paragraphs.

\vspace{0.05in}
\noindent{\bf Stage 1: Feature alignment.}
In this stage, we train a projector $g_P$ that maps image features into the textual embedding space of the LLM. This projector pre-training contributes to the stable vision-language alignment. Following the training setup in multi-modal large language models~\cite{liu2023visual, liu2024improved}, we freeze the parameters of both the visual encoder $f_V$ and the language model $f_L$, and optimize only the projector $g_P$. 

\vspace{0.05in}
\noindent{\bf Stage 2: Visual instruction tuning.}
Following the projector alignment in Stage 1, this stage extends the alignment to the LLM. We freeze the visual encoder $f_V$ and fine-tune the projector $g_P$ and the language model $f_L$ using our multi-view VQA data, combined with the single-view visual instruction data from LLaVA~\cite{liu2023visual}. This step enables $f_L$ and $g_P$ to handle multi-view visual questions. We provide details of proposed multi-view VQA data in \Cref{sec:method:alignment}.

\begin{wrapfigure}{r}{0.40\textwidth}
    \vspace{-0.5cm}
  \centering
  \centerline{\includegraphics[width=0.92\linewidth]{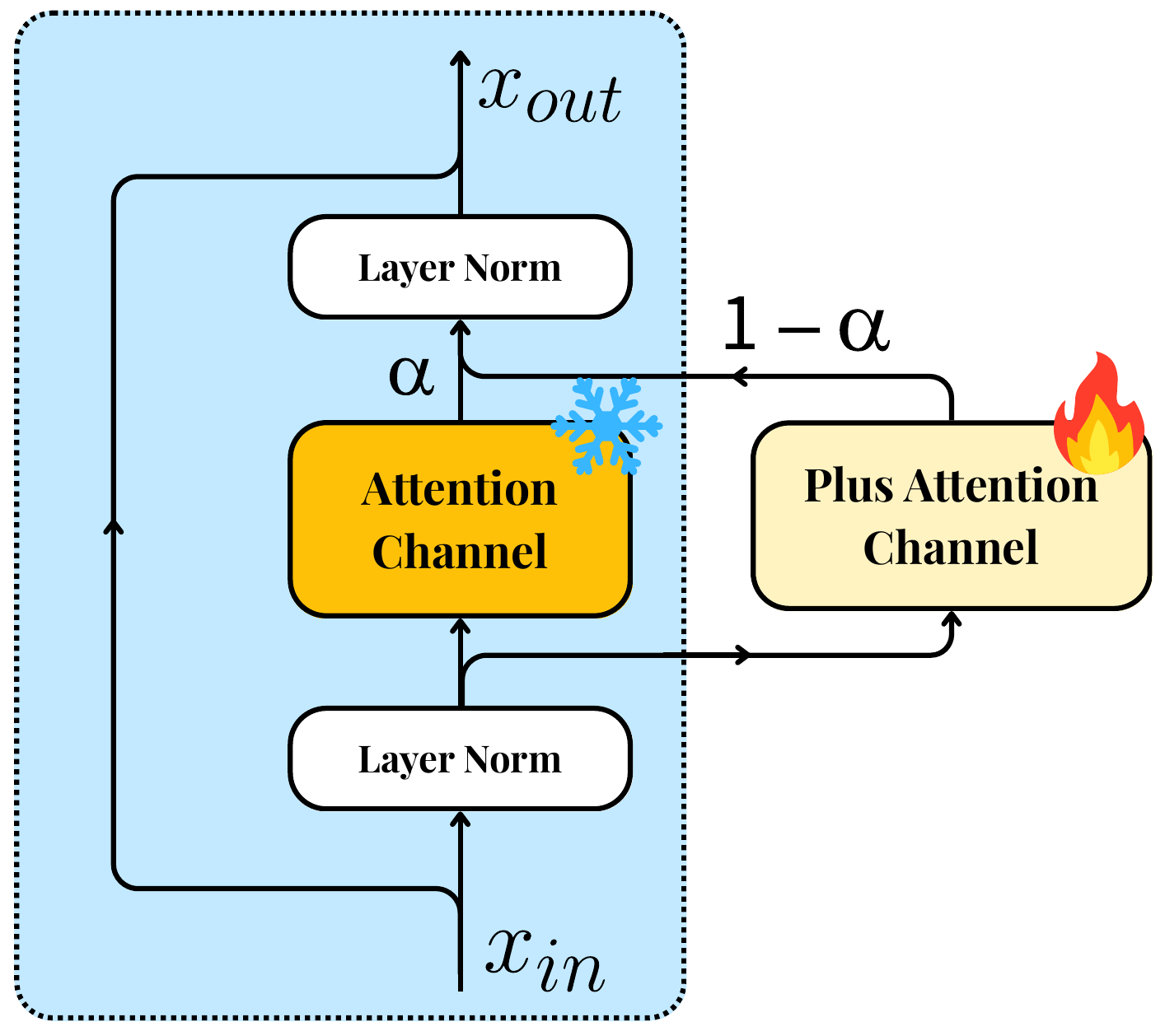}}
  \vspace{-0.10in}
    \caption{\textbf{Illustration of the dual-channel attention layer}~\cite{hong2022cogvideo}, where an additional attention block is introduced alongside the original attention block and merged via a learnable mixture factor $\alpha$.}
  \label{fig:dual_channel_att}
\vspace{-0.7cm}
\end{wrapfigure}
\vspace{0.05in}
\noindent{\bf Stage 3: Vision encoder fine-tuning with dual-channel attention.}
Finally, we fine-tune the vision encoder $f_V$ to have the capability of spatial understanding. To effectively inject dense spatial knowledge into the vision encoder, we use multi-turn visual spatial reasoning dataset (see \Cref{sec:method:dataset}), which is carefully designed for hierarchical spatial reasoning. We train the vision encoder $f_V$ and the projection module $g_P$ while keeping the parameters of the LLM $f_L$ frozen, allowing only the vision encoder to benefit from language-driven spatial information. We employ SFT loss, and through this training process, the vision encoder learns to extract meaningful representations necessary for producing answers. However, direct full fine-tuning risks forgetting of the pre-trained knowledge embedded in the vision encoder. To address this challenge, we introduce a dual-channel attention mechanism (see Figure \ref{fig:dual_channel_att}). Specifically, for each attention layer $\texttt{Attn}(\cdot)$ in the visual encoder $f_V$, we introduce an additional attention layer $\texttt{Attn}^{+}(\cdot)$, whose weight parameters are initialized to the same values as those of $\texttt{Attn}(\cdot)$. Given an input $\rvx$ to each attention layer, we merge the outputs of $\texttt{Attn}(\cdot)$ and $\texttt{Attn}^{+}(\cdot)$ by introducing a trainable mixture factor $\boldsymbol{\alpha} = \text{sigmoid}(\rva) \in (0, 1)^d$ with zero-initialized parameter $\rva \in \mathbb{R}^{d}$, where $d$ is the hidden dimension of $\rvx$, as follows:
\begin{equation}
\label{eq:dual}
\texttt{Attn}^{\mathrm{final}}(\rvx)
= \boldsymbol{\alpha} \cdot \texttt{Attn}(\rvx) +  (1-\boldsymbol{\alpha}) \cdot \texttt{Attn}^{+}(\rvx).
\end{equation}
During fine-tuning, we only update the parameters of $\texttt{Attn}^{+}$ and $\boldsymbol{\alpha}$ while keeping all other parameters frozen. This approach allows the vision encoder to initially rely on pre-trained attention weights and gradually incorporate new attention weights, smoothly enhancing spatial awareness without discarding existing knowledge (see classification result in \Cref{fig:ablation_class}).

\begin{figure*}[t]
\begin{minipage}{0.47\textwidth}
\centering\small
\captionof{table}{\textbf{Results on monocular depth estimation} from NYUd~\cite{silberman2012indoor} and KITTI~\cite{geiger2013vision} benchmarks. We report the RMSE score between ground truth and predicted depth values. Lower is better. For all results, we freeze the encoder backbone and train a linear head (lin.) or DPT head~\cite{ranftl2021vision} on top of the image features of the last layer.}
\vspace{.14in}
\resizebox{\linewidth}{!}{
\begin{tabular}{lcccc}
\toprule
& \multicolumn{2}{c}{NYUd} & \multicolumn{2}{c}{KITTI} \\
\cmidrule(lr){2-3} \cmidrule(lr){4-5}
Method & lin. & DPT & lin. & DPT \\
\midrule
\multicolumn{5}{l}{\textit{Vision-only trained encoder}} \\
V-JEPAv2~\cite{assran2025vjepa2} & 0.44 & 0.42 & 3.07 & 2.59 \\
\midrule
\multicolumn{5}{l}{\textit{Vision-Language trained encoder}} \\
AIMv2~\cite{fini2025aimv2} & 0.33 & 0.28 & 2.56 & 2.10 \\
\texttt{dino.txt}~\cite{jose2025dinomeetstext} & 0.42 & 0.39 & 2.98 & 2.65 \\
TIPS~\cite{maninis2025tips} & 0.35 & 0.31 & 2.58 & 2.11 \\
PE-Core~\cite{bolya2025perception} & 0.25 & 0.23 & 2.31 & 2.00 \\
\midrule
OpenCLIP & 0.53 & 0.41 & 3.54 & 2.70 \\
\textbf{+\sname (ours)} & \textbf{0.40} & \textbf{0.38} & \textbf{2.79} & \textbf{2.54} \\
SigLIPv2 & 0.51 & 0.40 & 3.32 & 2.64 \\
\textbf{+\sname (ours)} & \textbf{0.39} & \textbf{0.34} & \textbf{2.71} & \textbf{2.50} \\
\midrule
DINOv2 & 0.37 & 0.29 & 2.60 & 2.11 \\
\textbf{+\sname (ours)} & \textbf{0.30} & \textbf{0.25} & \textbf{2.53} & \textbf{2.07} \\
DINOv3 & 0.31 & 0.25 & 2.33 & 2.02 \\
\textbf{+\sname (ours)} & \textbf{0.25} & \textbf{0.21} & \textbf{2.20} & \textbf{1.84} \\
\bottomrule
\end{tabular}
}
\vspace{-.03in}
\label{tab:metric_depth}
\end{minipage}
\hfill
\begin{minipage}{0.48\textwidth}
\centering\small
\captionof{table}{\textbf{Results on semantic segmentation} from ADE20K~\cite{zhou2017scene} and Pascal VOC~\cite{Everingham10} benchmarks. We report mIoU score. Higher is better. For all results, we freeze the encoder backbone and report results of linear probing (lin.) or multi-scale evaluation (+ms), where the multi-scale approach uses features from the last four layers of the visual encoder to perform segmentation.}
\resizebox{\linewidth}{!}{
\begin{tabular}{lcccc}
\toprule
& \multicolumn{2}{c}{ADE20K} & \multicolumn{2}{c}{Pascal VOC} \\
\cmidrule(lr){2-3} \cmidrule(lr){4-5}
Method & lin. & +ms & lin. & +ms \\
\midrule
\multicolumn{5}{l}{\textit{Vision-only trained encoder}} \\
V-JEPAv2~\cite{assran2025vjepa2} & 51.3 & 53.8 & 83.2 & 86.6 \\
\midrule
\multicolumn{5}{l}{\textit{Vision-Language trained encoder}} \\
AIMv2~\cite{fini2025aimv2} & 31.9 & 37.9 & 66.6 & 72.1 \\
\texttt{dino.txt}~\cite{jose2025dinomeetstext} & 50.6 & 52.8 & 83.4 & 86.3 \\
TIPS~\cite{maninis2025tips} & 49.9 & 54.1 & 83.6 & 86.7 \\
PE-Core~\cite{bolya2025perception} & 41.5 & 48.0 & 73.2 & 80.5 \\
\midrule
OpenCLIP & 39.5 & 46.0 & 71.7 & 79.3 \\
\textbf{+\sname (ours)} & \textbf{40.5} & \textbf{47.3} & \textbf{75.1} & \textbf{80.9} \\
SigLIPv2 & 42.8 & 48.7 & 72.6 & 79.1 \\
\textbf{+\sname (ours)} & \textbf{45.1} & \textbf{50.8} & \textbf{79.0} & \textbf{82.2} \\
\midrule
DINOv2 & 49.3 & 53.0 & 83.0 & 86.2 \\
\textbf{+\sname (ours)} & \textbf{52.0} & \textbf{54.9} & \textbf{84.5} & \textbf{87.6} \\
DINOv3 & 55.9 & 60.3 & 86.6 & 89.8 \\
\textbf{+\sname (ours)} & \textbf{59.7} & \textbf{63.1} & \textbf{88.5} & \textbf{90.9} \\
\bottomrule               
\end{tabular}}
\label{tab:semantic_seg}
\end{minipage}
\vspace{-0.4cm}
\end{figure*}

\subsection{Enhancing Vision Encoder with Spatial CoT}\label{sec:method:CoT_data}
To effectively inject dense spatial information into vision encoders, we address the fundamental limitations of existing spatial datasets. Current spatial VQA data consist of simple single-turn QA pairs with limited information content, insufficient for transferring comprehensive 3D understanding. To overcome this limitation, We introduce Multi-view VQA, which helps align the vision encoder with the LLM to effectively handle multi-view data and a multi-turn Chain-of-Thought (CoT) framework~\cite{wei2022chain} for both single-view and multi-view images that enables the injection of substantially richer spatial information in a single training instance.

\begin{table*}[t]
\centering
\small
\caption{\textbf{Results on 3D-centric tasks.} We evaluate unified probing on diverse 3D-related tasks from ScanNet~\cite{dai2017scannet} scenes. We report BLEU-1 score for Vision-Language Reasoning (VLR) on ScanQA~\cite{azuma2022scanqa} and SQA3D~\cite{ma2022sqa3d}. For Visual Grounding (VG), we report accuracy (\%) on overall category of ScanRefer~\cite{chen2020scanrefer} dataset. For Geometric Understanding (GU), we report Registration Recall (RR) at 0.05m RMSE threshold and Relative Translation Error (RTE). For 3D Semantic Understanding (3D SU), we report accuracy and mIoU. Lower is better for RTE and higher is better for all other metrics.}
\vspace{-0.08in}
\setlength{\tabcolsep}{6pt}
\resizebox{\textwidth}{!}{
\begin{tabular}{lccccccc}
\toprule
 & \multicolumn{2}{c}{VLR} & VG & \multicolumn{2}{c}{GU} & \multicolumn{2}{c}{3D SU}  \\
\cmidrule(lr){2-3} \cmidrule(lr){4-4} \cmidrule(lr){5-6} \cmidrule(lr){7-8}
Method & ScanQA $\uparrow$ & SQA3D $\uparrow$ & ScanRefer (\%) $\uparrow$ & RR@0.05m (\%) $\uparrow$ & RTE (m) $\downarrow$ & Acc $\uparrow$ & mIoU $\uparrow$ \\
\midrule
\multicolumn{8}{l}{\textit{Vision-only trained encoder}} \\
V-JEPAv2~\cite{assran2025vjepa2} & 37.7 & 49.0 & 55.5 & 95.4 & 0.08 & 69.5 & 33.2 \\
\midrule
\multicolumn{8}{l}{\textit{Vision-Language trained encoder}} \\
AIMv2~\cite{fini2025aimv2} & 37.4 & 48.1 & 50.9 & 47.8 & 0.31 & 53.9 & 11.8 \\
\texttt{dino.txt}~\cite{jose2025dinomeetstext} & 39.8 & 50.4 & 52.5 & 81.7 & 0.16 & 85.7 & 59.4 \\
TIPS~\cite{maninis2025tips} & 37.4 & 49.2 & 52.0 & 84.0 & 0.16 & 71.3 & 39.7 \\
PE-Core~\cite{bolya2025perception} & 40.5 & 51.7 & 51.7 & 44.6 & 0.29 & 85.0 & 47.5 \\
\midrule
OpenCLIP & 36.9 & 48.0 & 50.1 & 22.6 & 0.40 & 39.8 & 6.9 \\
\textbf{+\sname (ours)} & \textbf{39.2} & \textbf{49.9} & \textbf{56.6} & \textbf{78.8} & \textbf{0.17} & \textbf{76.9} & \textbf{54.9} \\
SigLIPv2 & 38.1 & 48.5 & 51.4 & 47.8 & 0.28 & 47.7 & 9.2 \\
\textbf{+\sname (ours)} & \textbf{40.8} & \textbf{50.1} & \textbf{56.8} & \textbf{86.4} & \textbf{0.15} & \textbf{81.0} & \textbf{55.5} \\
\midrule
DINOv2 & 39.5 & 49.8 & 52.7 & 82.4 & 0.15 & 83.0 & 64.1 \\
\textbf{+\sname (ours)} & \textbf{40.3} & \textbf{50.4} & \textbf{57.0} & \textbf{92.4} & \textbf{0.13} & \textbf{89.8} & \textbf{68.3} \\
DINOv3 & 40.6 & 51.4 & 56.2 & 86.9 & 0.10 & 91.1 & 69.1 \\
\textbf{+\sname (ours)} & \textbf{43.3} & \textbf{54.9} & \textbf{61.1} & \textbf{97.5} & \textbf{0.06} & \textbf{91.9} & \textbf{70.6} \\
\bottomrule
\end{tabular}}
\label{tab:lexicon3d}
\vspace{-0.4cm}
\end{table*}
\vspace{0.05in}
\noindent{\bf Multi-view VQA Dataset.} \label{sec:method:alignment} To enhance multi-view VQA capabilities during the visual instruction tuning (Stage 2), we construct multi-view VQA dataset. We first apply LPIPS~\cite{zhang2018unreasonable} metric to the 3D or video dataset to obtain a pair of images. Given the pair of images, we employ GPT-4o~\cite{achiam2023gpt} to generate visual questions targeting general multi-view knowledge, which do not require spatial knowledge. We provide more details in \Cref{appendix:multiview_vqa}.

\vspace{0.05in}
\noindent{\bf Multi-turn Visual Spatial Reasoning Dataset.}
\label{sec:method:dataset}
To enhance spatial reasoning capabilities of the vision encoder (Stage 3), we construct multi-turn visual spatial reasoning dataset for single-view and multi-view. Additionally, to enhance general knowledge of the vision encoder, we append GPT-generated scene captions after spatial reasoning turn.
For single-view image, we first extract a 3D point cloud from given an image $\rvx$ by applying diverse vision models (\eg, depth estimation model~\cite{bochkovskii2024depth} and image segmentation model~\cite{ravi2024sam2segmentimages}). For multi-view images $\{ \rvx_{\texttt{1}}, \cdots, \rvx_{\texttt{N}} \}$, we use 3D reconstruction model~\cite{wang2025vggt} to extract a 3D point cloud from given images. Using the point cloud, we synthesize QA pairs specialized in spatial reasoning about $\rvx$ or $\{ \rvx_{\texttt{1}}, \cdots, \rvx_{\texttt{N}} \}$.

We then design spatial reasoning QA pairs at three hierarchical levels: pixel, object, and scene, enabling LLM to perform CoT reasoning from narrow to broad view.
Specifically, at the pixel-level, the QA task is designed to capture the overall geometry in the image by querying the absolute or relative 3D position of a point, \eg, ``What is the depth value at coordinate $(x, y)$?".
At the object-level, the QA task tackles the semantic spatial information of objects inside the image using a bounding cube of the object in 3D space, \eg, 
``Is [A] on the left side of [B]?", where [A] and [B] is the descriptions about the object in image. We note that this level uses the pixel-level spatial information as a rationale, enabling LLM to reason about the geometry of objects in 3D space. Lastly, at the scene-level, the QA task is designed to predict the exact distance between multiple objects that requires coherent 3D spatial understanding, \eg, ``How far is [A] from [B]?".
\section{Experiments}
\begin{figure*}[t]
\begin{minipage}{0.66\textwidth}
\centering\small
\captionof{table}{
\textbf{Results on vision-based robot learning.} We report the performance of imitation learning agents on 4 domains from CortexBench~\cite{majumdar2023we}, which are trained upon the image representations. In particular, we report the normalized score for DMControl and success rates (\%) for other tasks.}
\begin{adjustbox}{max width=\linewidth}
\begin{tabular}{lccccc}
\toprule
Method & Adroit & MetaWorld & DMControl & Trifinger & Avg. \\
\midrule
OpenCLIP & 52.6 $\pm$ 4.9 & 83.0 $\pm$ 2.7 & 58.5 $\pm$ 1.9 & 67.7 $\pm$ 0.5 & 65.5 \\
\textbf{+\sname (ours)} & \textbf{61.1} $\pm$ 3.4 & \textbf{87.0} $\pm$ 3.3 & \textbf{61.0} $\pm$ 1.6 & \textbf{72.9} $\pm$ 0.3 & \textbf{70.5} \\
\midrule
SigLIPv2 & 56.5 $\pm$ 3.0 & 84.7 $\pm$ 2.9 & 69.4 $\pm$ 2.1 & 68.3 $\pm$ 0.8 & 69.7 \\
\textbf{+\sname (ours)} & \textbf{66.5} $\pm$ 1.9 & \textbf{89.1} $\pm$ 0.9 & \textbf{73.5} $\pm$ 1.8 & \textbf{73.9} $\pm$ 0.7 & \textbf{75.8} \\
\midrule
DINOv2 & 55.4 $\pm$ 2.7 & 82.4 $\pm$ 4.0 & 67.9 $\pm$ 1.0 & 66.8 $\pm$ 0.2 & 68.1 \\
\textbf{+\sname (ours)} & \textbf{68.1} $\pm$ 2.9 & \textbf{88.5} $\pm$ 3.1 & \textbf{75.0} $\pm$ 1.1 & \textbf{71.4} $\pm$ 0.8 & \textbf{75.8} \\
\midrule
DINOv3 & 63.9 $\pm$ 1.5 & 83.8 $\pm$ 1.6 & 70.8 $\pm$ 1.8 & 72.8 $\pm$ 0.5 & 72.8 \\
\textbf{+\sname (ours)} & \textbf{71.8} $\pm$ 3.4 & \textbf{92.0} $\pm$ 1.9 & \textbf{80.4} $\pm$ 2.4 & \textbf{79.0} $\pm$ 0.6 & \textbf{80.8} \\
\bottomrule
\end{tabular}
\end{adjustbox}
\label{tab:cortex_bench}
\vspace{+0.12in}
\end{minipage}
\hfill
\begin{minipage}{0.32\textwidth}
\centering
\centerline{\includegraphics[width=0.7\linewidth]{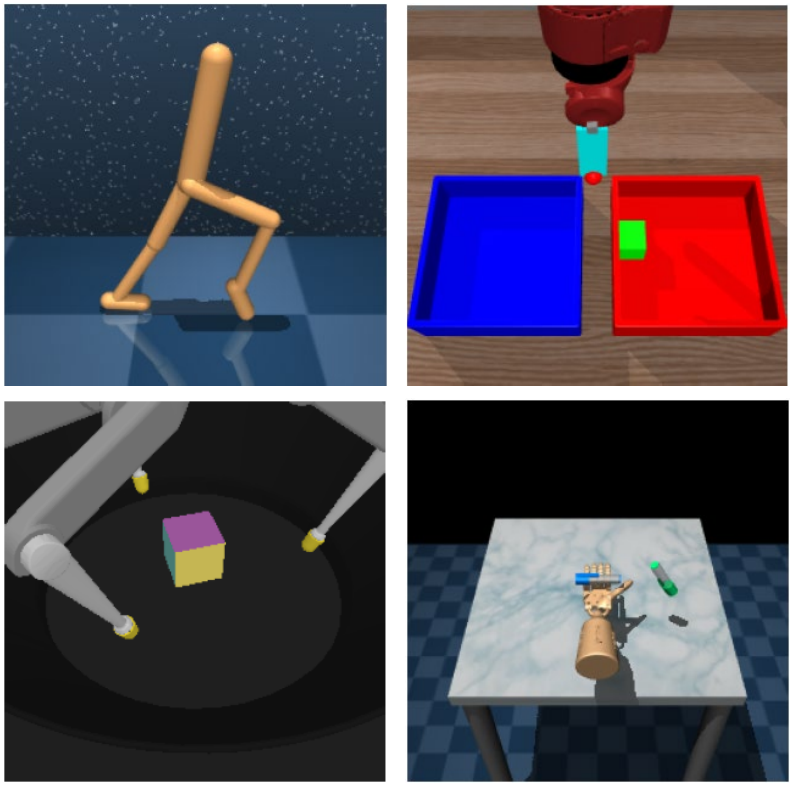}}
\vspace{-0.08in}
\caption{\textbf{Examples of visual observations from CortexBench.} We train imitation learning agents to learn a mapping from these visual observations to expert actions.
}
\label{fig:cortexbench_example}
\end{minipage}
\vspace{-0.4cm}
\end{figure*}

Through extensive experiments, we validate the performance of \sname and ablate its key components, focusing on following questions:
\vspace{-0.02in}
\begin{itemize}[leftmargin=*,itemsep=1mm, label=\textbullet]
    \item Can \sname improve spatial knowledge of the vision encoder?
    
    (\Cref{tab:metric_depth,tab:semantic_seg,tab:cortex_bench,tab:lexicon3d})
    \item Isn't \sname overfitted to spatial knowledge? (\Cref{tab:classification})
    \item Which components contribute to \sname performance? 
    
    (Tables~\ref{tab:analysis_curriculum} to \ref{tab:analysis_component} and \Cref{fig:ablation_class})
\end{itemize}

\subsection{Experimental Setup}

\vspace{0.05in}
\noindent{\bf VQA Dataset Construction.}
For single-view image, we use randomly sampled 100K images from the SA1B dataset~\cite{kirillov2023segment} to construct the single-view VQA dataset specialized in chain-of-thought spatial reasoning. For multi-view images, we use filtered 200K samples from the ego-centric video dataset~\cite{grauman2022ego4d} and 3D dataset~\cite{jensen2014dtu, dai2017scannet, mildenhall2021nerf, barron2022mip} to construct multi-view VQA dataset niche in multi-view reasoning or alignment. More details are provided in \Cref{appendix:implementation_details_data}.

\vspace{0.05in}
\noindent{\bf Baselines.}
For all experiments, we compare our methods with the recent widely-used pre-trained image representation models. 
To be specific, we first consider OpenCLIP~\cite{cherti2023reproducible} ViT-G/14 and SigLIPv2~\cite{tschannen2025siglip2} ViT-g/16, known for language-aligned vision encoder. We also consider DINOv2~\cite{oquab2023dinov2} ViT-g/14 and DINOv3~\cite{simeoni2025dinov3} ViT-7B/16, which is a recent state-of-the-art vision encoder. We further include comparable methods, including the vision-only trained encoder like V-JEPAv2~\cite{assran2025vjepa2}, as well as the vision-language trained encoders such as AIMv2~\cite{fini2025aimv2}, \texttt{dino.txt}~\cite{jose2025dinomeetstext}, TIPS~\cite{maninis2025tips}, and Perception Encoder~\cite{bolya2025perception}.

\vspace{0.05in}
\noindent{\bf Implementation Details.}
We choose Qwen-2.0-7B~\cite{yang2024qwen2} as the LLM backbone and 2-layer MLP as the projector, following the architecture of LLaVA-1.5~\cite{liu2024improved}. Further details are provided in \Cref{appendix:implementation_details}.

\begin{table*}[t]
\centering
\small
\caption{\textbf{Results on image classification and retrieval tasks.} We report Top-1 accuracy of kNN performance and linear probing (lin.) for image classification on validation set of ImageNet-1K~\cite{russakovsky2015imagenet}. For image retrieval, we report global average precision (GAP) on Met~\cite{ypsilantis2021met} and mean average precision (mAP) on Oxford-Hard (Oxford-H)~\cite{radenovic2018revisiting}, Paris-Hard (Paris-H)~\cite{radenovic2018revisiting}, and AmsterTime dataset~\cite{yildiz2022amstertime}. For all results, we freeze the encoder backbone.}
\vspace{-0.08in}
\setlength{\tabcolsep}{6pt}
\resizebox{\textwidth}{!}{
\begin{tabular}{lcccccc}
\toprule
 & \multicolumn{2}{c}{Image classification} & \multicolumn{4}{c}{Image retrieval} \\
\cmidrule(lr){2-3} \cmidrule(lr){4-7}
Method & ImageNet (kNN) & ImageNet (lin.) & Oxford-H & Paris-H & Met (GAP) & AmsterTime \\
\midrule
\multicolumn{7}{l}{\textit{Vision-only trained encoder}} \\
V-JEPAv2~\cite{assran2025vjepa2} & 82.1 & 84.0 & 40.6 & 76.1 & 38.2 & 31.9 \\
\midrule
\multicolumn{7}{l}{\textit{Vision-Language trained encoder}} \\
AIMv2~\cite{fini2025aimv2} & 85.7 & 88.3 & 60.8 & 86.4  & 52.0 & 50.3 \\
\texttt{dino.txt}~\cite{jose2025dinomeetstext} & 80.0 & 81.4 & 49.7 & 78.5 & 41.8 & 33.1 \\
TIPS~\cite{maninis2025tips} & 83.3 & 86.2 & 29.0 & 60.0 & 20.1 & 18.5 \\
PE-Core~\cite{bolya2025perception} & 86.8 & 89.5 & 27.9 & 64.1 & 15.9 & 20.0 \\
\midrule
OpenCLIP & 84.0 & 86.8 & 23.4 & 59.7 & 7.4 & 24.4 \\
\textbf{+\sname (ours)} & \textbf{86.1} & \textbf{87.9} & \textbf{32.8} & \textbf{69.4} & \textbf{19.7} & \textbf{30.3} \\
SigLIPv2 & 86.3 & 89.1 & 25.1 & 60.9 & 13.9 & 15.5 \\
\textbf{+\sname (ours)} & \textbf{87.6} & \textbf{90.0} & \textbf{36.0} & \textbf{69.1} & \textbf{24.0} & \textbf{27.2} \\
\midrule
DINOv2 & 84.5 & 87.3 & 58.2 & 84.6 & 44.6 & 48.9 \\
\textbf{+\sname (ours)} & \textbf{86.4} & \textbf{88.6} & \textbf{61.3} & \textbf{85.2} & \textbf{45.1} & \textbf{50.8} \\
DINOv3 & 85.8 & 88.4 & 60.7 & 87.1 & 55.4 & 56.5 \\
\textbf{+\sname (ours)} & \textbf{87.7} & \textbf{90.2} & \textbf{64.1} & \textbf{88.6} & \textbf{57.0} & \textbf{56.9} \\
\bottomrule
\end{tabular}}
\label{tab:classification}
\end{table*}
\subsection{Dense Prediction Tasks}\label{sec:exp:dense_prediction}

\vspace{0.05in}
\noindent{\bf Setup.}
We evaluate \sname on dense prediction tasks requiring geometric and semantic spatial understanding. For geometric understanding, we perform monocular depth estimation on NYUd~\cite{silberman2012indoor} and KITTI~\cite{geiger2013vision} using linear or DPT~\cite{ranftl2021vision} heads. For semantic understanding, we evaluate on ADE20K~\cite{zhou2017scene} and Pascal VOC~\cite{Everingham10} segmentation benchmarks using linear or multi-scale heads. All experiments freeze the visual backbone during training (see \Cref{appendix:implementation_details} for details).

\vspace{0.05in}
\noindent{\bf Results.}
As shown in Table~\ref{tab:metric_depth} and \ref{tab:semantic_seg}, \sname consistently improves both geometric and semantic spatial understanding across various encoders. For instance, OpenCLIP's RMSE on NYUd decreases from 0.53 to 0.40 with a linear head, while DINOv3's mIoU on ADE20K increases from 55.9\% to 59.7\%. These consistent gains demonstrate that language-based spatial knowledge transfer effectively enhances visual encoders' spatial understanding capabilities.

\subsection{Complex 3D-centric Tasks}
\label{sec:exp:3d}

\vspace{0.05in}
\noindent{\bf Setup.}
We evaluate \sname on Lexicon3D~\cite{man2024lexicon3d}, a unified benchmark for 3D scene understanding covering vision-language reasoning, visual grounding, semantic understanding, and geometric understanding. Following Lexicon3D protocols, we freeze visual backbones and train task-specific heads (see \Cref{appendix:implementation_details} for details).

\vspace{0.05in}
\noindent{\bf Results.}
As shown in \Cref{tab:lexicon3d}, \sname shows comprehensive improvements across diverse 3D tasks. OpenCLIP's BLEU-1 improves from 36.9 to 39.2 on ScanQA~\cite{azuma2022scanqa}, while DINOv3 increases from 51.4 to 54.9 on SQA3D~\cite{ma2022sqa3d}, demonstrating that \sname improves spatial understanding without compromising language capabilities. Notably, OpenCLIP's 3D semantic segmentation dramatically improves from 6.9 to 54.9 mIoU, highlighting \sname can inject robust spatial knowledge into encoders with initially limited spatial awareness.

\begin{table*}[t]
\centering
\small
\captionof{table}{
\textbf{Effect of LLM-based fine-tuning.} We fine-tune the vision encoder with different headers. We report accuracy (\%) for classification (Cls) on ImageNet-1K, mIoU for segmentation (Seg) on ADE20K, RMSE for depth estimation on NYUd, and BLEU-1 score for vision-language reasoning (VLR) on ScanQA. We use ViT-L/14 as the backbone architecture of the encoder.}
\label{tab:analysis_curriculum}
\vspace{-0.08in}
\begin{adjustbox}{max width=0.8\linewidth}
\begin{tabular}{lccccc}
\toprule
Method & Cls $\uparrow$ & Seg $\uparrow$ & Depth $\downarrow$ & VLR $\uparrow$\\
\midrule
DINOv2 & 86.3 & 47.7 & 0.38 & 39.2 \\
\midrule
+Linear (depth) & 85.7 (\textcolor{red!60!black}{-1.39\%}) & 47.9 (\textcolor{green!60!black}{+0.42\%}) & 0.35 (\textcolor{green!60!black}{-7.89\%}) & 36.9 (\textcolor{red!60!black}{-5.87\%}) \\
+Linear (seg.)  & 86.6 (\textcolor{green!60!black}{+0.35\%}) & 48.8 (\textcolor{green!60!black}{+2.31\%}) & 0.45 (\textcolor{red!60!black}{+18.42\%}) & 37.1 (\textcolor{red!60!black}{-5.36\%}) \\
+SAM decoder     & 86.3 (\textcolor{green!60!black}{+0.0\%}) & 50.1 (\textcolor{green!60!black}{+5.03\%}) & 0.42 (\textcolor{red!60!black}{+10.53\%}) & 37.6 (\textcolor{red!60!black}{-4.08\%})\\
+VGGT decoder    & 84.8 (\textcolor{red!60!black}{-1.74\%}) & 45.6 (\textcolor{red!60!black}{-4.40\%}) & 0.35 (\textcolor{green!60!black}{-7.89\%}) & 37.3 (\textcolor{red!60!black}{-4.85\%}) \\
\rowcolor{skyblue}
+LLM (ours)           & 88.3 (\textcolor{green!60!black}{+2.32\%}) & 51.5 (\textcolor{green!60!black}{+7.97\%}) & 0.32 (\textcolor{green!60!black}{-15.79\%}) & 40.0 (\textcolor{green!60!black}{+2.04\%}) \\
\bottomrule
\end{tabular}
\end{adjustbox}
\vspace{-0.2cm}
\end{table*}
\begin{figure*}[t]
  \centering
  \begin{subfigure}{.32\linewidth}
    \centering
    \includegraphics[width=\linewidth]{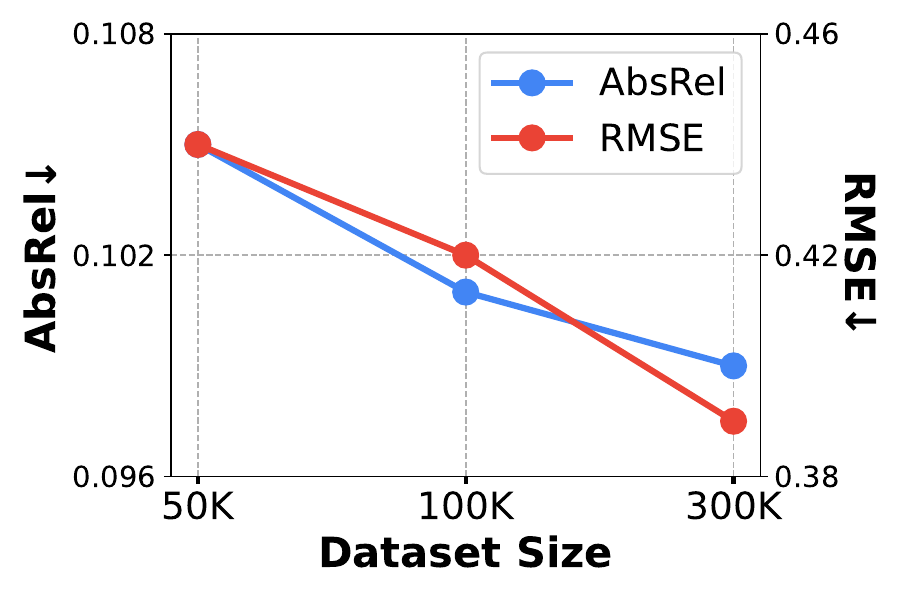}
    \caption{SigLIPv2 depth estimation}
  \end{subfigure}
  \hfill
  \begin{subfigure}{.32\linewidth}
    \centering
    \includegraphics[width=\linewidth]{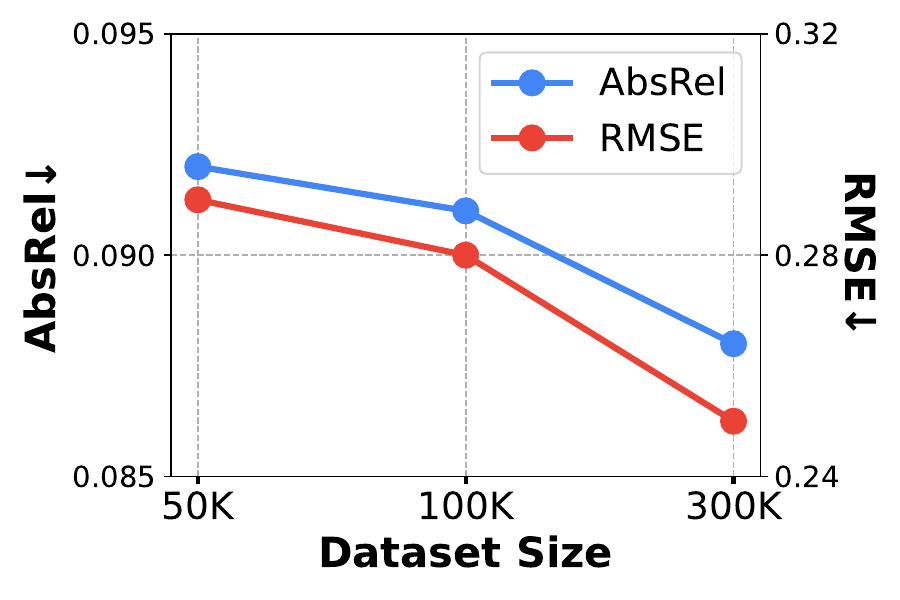}
    \caption{DINOv3 depth estimation}
  \end{subfigure}
  \hfill
  \begin{subfigure}{.32\linewidth}
    \centering
    \includegraphics[width=\linewidth]{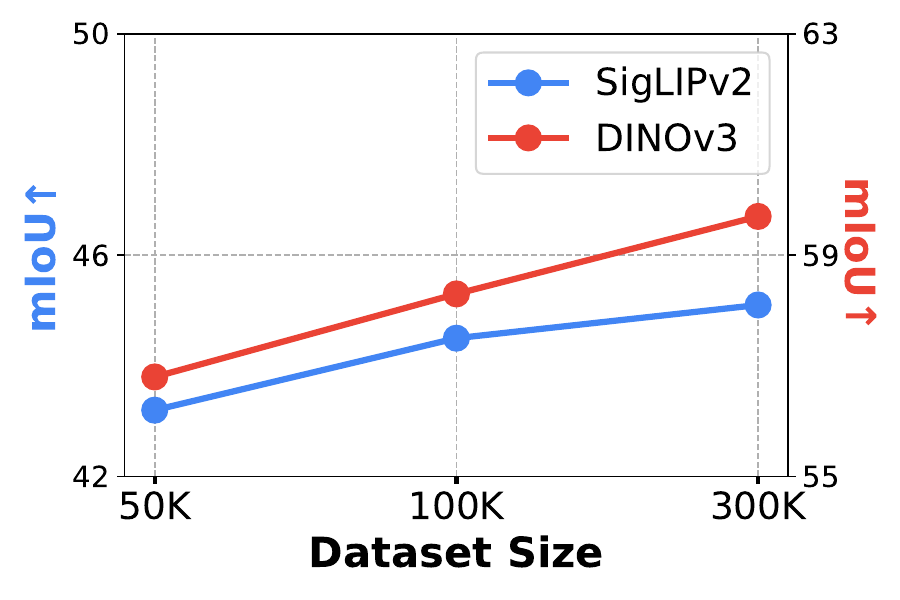}
    \caption{Semantic segmentation}
  \end{subfigure}
  \caption{\textbf{Effect of dataset scalability.} We investigate the effect of the size of analysis of data scalability effects on (a) depth estimation results (AbsRel, RMSE) on NYUd benchmark for SigLIPv2, (b) depth estimation results (AbsRel, RMSE) on NYUd benchmark for DINOv3, and (c) semantic segmentation results (mIoU) on ADE20K benchmark for SigLIPv2 and DINOv3. The results show scalable performance improvements with increased data size.}
  \label{fig:ablation_datasize}
  \vspace{-0.4cm}
\end{figure*}
\subsection{Vision-based Robot Learning}\label{sec:exp:robot}

\vspace{0.05in}
\noindent{\bf Setup.}
We evaluate \sname on vision-based robot control using 4 domains from CortexBench~\cite{majumdar2023we} spanning locomotion and manipulation tasks~\cite{rajeswaran2017learning,yu2020meta,tassa2018deepmind,wuthrich2020trifinger}. Following CortexBench protocols, we train behavior cloning agents using \texttt{[CLS]} representations to predict expert actions from visual observations. We report the mean of best performance across 5 evaluation runs (see \Cref{appendix:implementation_details} for details).

\vspace{0.05in}
\noindent{\bf Results.}
As shown in \Cref{tab:cortex_bench}, \sname significantly improves robot task performance across all vision encoders. For example, DINOv2 + \sname achieves 68.1\% on Adroit versus 55.4\% for DINOv2 alone, demonstrating that enhanced spatial representations directly benefit robot control.
\vspace{-0.1cm}

\subsection{Image Classification and Retrieval Tasks}\label{sec:exp:classification}

\vspace{0.05in}
\noindent{\bf Setup.}
We evaluate \sname's impact on instance recognition using ImageNet-1K~\cite{russakovsky2015imagenet} classification and retrieval benchmarks (Oxford, Paris~\cite{radenovic2018revisiting}, Met~\cite{ypsilantis2021met}, AmsterTime~\cite{yildiz2022amstertime}). Following DINOv3 protocols, we use linear probing on \texttt{[CLS]} representations for classification and similarity-based ranking for retrieval (see \Cref{appendix:implementation_details} for details).

\vspace{0.05in}
\noindent{\bf Results.}
As shown in \Cref{tab:classification}, \sname improves both classification and retrieval despite these tasks not explicitly requiring spatial understanding. DINOv3's ImageNet accuracy increases from 88.4\% to 90.2\%, while Oxford-Hard mAP improves from 60.7 to 64.1. These results demonstrate that \sname enhances general vision capabilities without overfitting to spatial features, likely due to our dual-channel attention preserving pre-trained knowledge and the inclusion of general scene captions alongside spatial reasoning.

\begin{table*}[t]
\centering
\small
\vspace{+0.08in}
\captionof{table}{
\textbf{Component-wise analysis.} We investigate the effect of multi-turn spatial reasoning data and the effect of single-view and multi-view data. Multi-turn order means the order of three levels (\ie, pixel, object, and scene) in our visual spatial reasoning data. We report accuracy (\%) for classification (Cls) on ImageNet-1K, mIoU for segmentation (Seg) on ADE20K, RMSE for depth estimation on NYUd.}
\label{tab:analysis_component}
\vspace{-0.08in}
\begin{adjustbox}{max width=0.9\linewidth}
    \begin{tabular}{lcccccc}
\toprule
Method & Multi-turn order & Single-view data & Multi-view data & Cls $\uparrow$ & Seg $\uparrow$ & Depth $\downarrow$ \\
\midrule
DINOv2 & \xmark & - & - & 86.3 & 47.7 & 0.38 \\
\midrule
+\sname           & Reverse & +100K & - & 87.4 & 48.4 & 0.35 \\
\phantom{+ Ours} & Random  & +100K & - & 87.4 & 48.5 & 0.36 \\
\phantom{+ Ours} & Forward & +100K & - & 87.6 & 48.9 & 0.34 \\ 
\cmidrule{2-7}
\phantom{+ Ours} & Forward & - & +100K & 87.6 & 48.2 & 0.36 \\
\phantom{+ Ours} & \cellcolor{skyblue}Forward & \cellcolor{skyblue}+50K & \cellcolor{skyblue}+50K & \cellcolor{skyblue}87.6 & \cellcolor{skyblue}49.2 & \cellcolor{skyblue}0.32 \\
\bottomrule
\end{tabular}
\end{adjustbox}
\vspace{-0.4cm}
\end{table*}
\subsection{Ablation Study and Analysis}\label{sec:exp:ablation}

\vspace{0.05in}
\noindent{\bf Effect of LLM-based Fine-tuning.} 
In \Cref{tab:analysis_curriculum}, we investigate whether LLM-based decoders provide superior supervision compared to pixel-level alternatives. We fine-tune the vision encoder with linear layer, SAM~\cite{kirillov2023segment} decoder, VGGT~\cite{wang2025vggt} decoder, and LLM~\cite{yang2024qwen2}. We then evaluate encoders on ImageNet-1K classification, ADE20K segmentation, and NYUd depth estimation. The results show that LLM consistently outperform pixel-level supervision methods, validating that language provides superior dense information transfer for vision encoders (see \Cref{appendix:detail_ablation} for details).

\vspace{0.05in}
\noindent{\bf Effect of Multi-turn Visual Reasoning.}
In \Cref{tab:analysis_component}, we investigate how the hierarchical structure of reasoning affects representation learning. We compare dataset construction strategies: (a) shuffled multi-turn, (b) reversed order (scene→object→pixel), and (c) forward order (pixel→object→scene). The forward hierarchical ordering shows optimal performance, demonstrating that reasoning order significantly impacts the quality of representation.

\begin{table*}[t]
\centering
\small
\captionof{table}{
\textbf{Comparison with post-training.} We fine-tune vision encoders with their original pre-training objectives (simple FT). We report RMSE for monocular depth estimation on NYUd, mIoU for semantic segmentation on ADE20K, BLEU-1 score for vision-language (VL) reasoning on ScanQA, average score for robot learning on CortexBench, and Top-1 accuracy (\%) for classification on ImageNet-1K. Lower is better for depth estimation.}
\label{tab:compare_pretrain}
\vspace{-0.08in}
\begin{adjustbox}{max width=\linewidth}
\begin{tabular}{lccccc}
\toprule
Method & Depth Estimation $\downarrow$ & Segmentation $\uparrow$ & VL Reasoning $\uparrow$ & Robot Learning $\uparrow$ & Classification $\uparrow$ \\
\midrule
OpenCLIP & 0.53 & 39.5 & 36.9 & 65.5 & 84.0 \\
+Simple FT & 0.56 & 39.6 & 37.7 & 63.7 & 84.3 \\
\rowcolor{skyblue}
+\sname (ours) & \textbf{0.40} & \textbf{40.5} & \textbf{39.2} & \textbf{72.9} & \textbf{86.1} \\
\midrule
SigLIPv2 & 0.51 & 42.8 & 38.1 & 69.7 & 86.3 \\
+Simple FT & 0.53 & 43.0 & 38.4 & 67.9 & 86.4 \\
\rowcolor{skyblue}
+\sname (ours) & \textbf{0.39} & \textbf{45.1} & \textbf{40.8} & \textbf{75.8} & \textbf{87.6} \\
\midrule
DINOv2 & 0.37 & 49.3 & 39.5 & 68.1 & 84.5 \\
+Simple FT & 0.36 & 49.6 & 39.4 & 69.4 & 84.7 \\
\rowcolor{skyblue}
+\sname (ours) & \textbf{0.30} & \textbf{52.0} & \textbf{40.3} & \textbf{75.8} & \textbf{86.4} \\
\midrule
DINOv3 & 0.31 & 55.9 & 40.6 & 72.8 & 85.8 \\
+Simple FT & 0.31 & 56.4 & 40.2 & 75.5 & 86.1 \\
\rowcolor{skyblue}
+\sname (ours) & \textbf{0.25} & \textbf{59.7} & \textbf{43.3} & \textbf{80.8} & \textbf{87.7} \\
\bottomrule
\end{tabular}
\end{adjustbox}
\vspace{-0.4cm}
\end{table*}

\vspace{0.05in}
\noindent{\bf Effect of Single-view and Multi-view Data.}
In \Cref{tab:analysis_component}, we investigate the effect of single-view and multi-view reasoning data. With fixed total samples, we compare single-view only, multi-view only, and combined training. While both data types independently improve performance, the  combination achieves the highest results, confirming their complementary nature.

\begin{wrapfigure}{r}{0.40\textwidth}
    \vspace{-0.8cm}
    \centering
    \centerline{\includegraphics[width=0.92\linewidth]{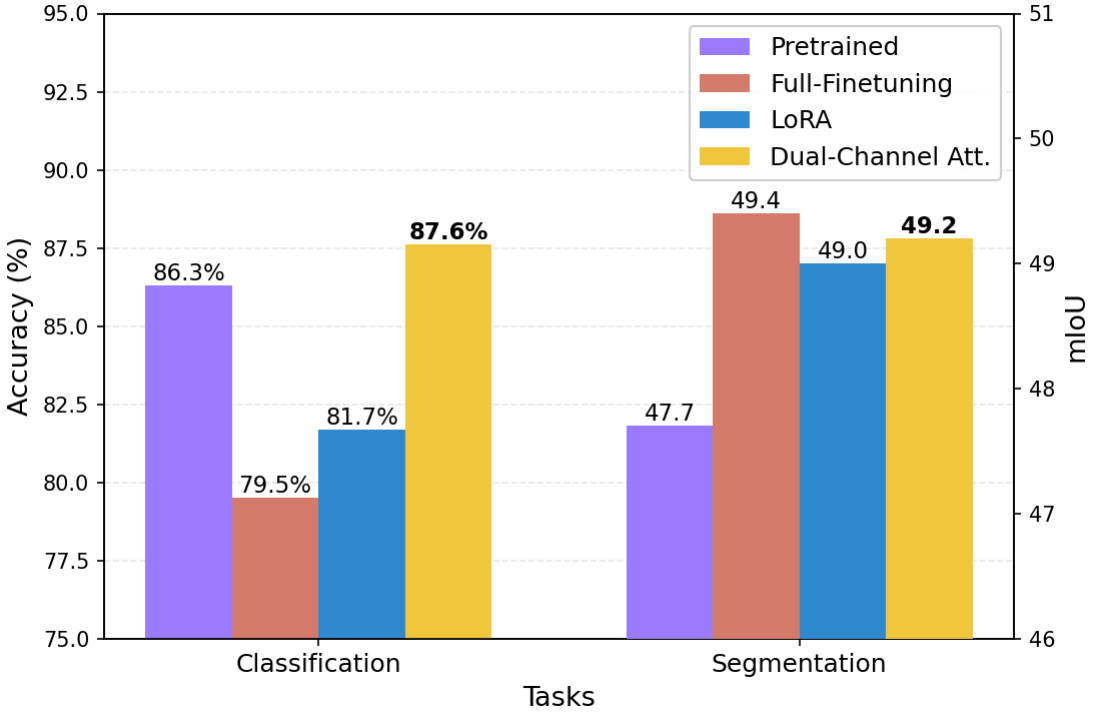}}
  \vspace{-0.10in}
    \caption{\textbf{Effect of dual-channel attention layer.} We report the linear evaluation performance of DINOv2-ViT-L/14 across different fine-tuning strategies.}
  \label{fig:ablation_class}
\vspace{-0.7cm}
\end{wrapfigure}

\vspace{0.05in}
\noindent{\bf Comparison with Naive Post-training.}
In \Cref{tab:compare_pretrain}, we investigate the effect of post-training. With fixed total samples (\ie, 300K data in multi-turn reasoning data), we compare the naive post-training scheme and \sname. We evaluate the performance of the vision encoder across five tasks: depth estimation, segmentation, vision-language reasoning, robot learning, and classification. 
The results show that naive post-training does not yield effective representations for downstream tasks.

\vspace{0.05in}
\noindent{\bf Effect of Dual-channel Attention Layer.}
In \Cref{fig:ablation_class}, we investigate whether our dual-channel attention mechanism preserves pre-trained knowledge during fine-tuning. We evaluate several approaches for fine-tuning the vision encoder including full fine-tuning, LoRA~\cite{hu2021loralowrankadaptationlarge}, and dual-channel~\cite{hong2022cogvideo} on ImageNet~\cite{russakovsky2015imagenet} and ADE20K~\cite{zhou2017scene}. Dual-channel attention uniquely preserves and even enhances pre-trained knowledge, while other approaches cause degradation.

\begin{table*}[t]
\centering
\small
\captionof{table}{
\textbf{Application to spatial-aware vision encoder.} We apply \sname on spatial-enhanced vision encoder~\cite{maninis2025tips, bolya2025perception}. We report RMSE for monocular depth estimation on NYUd, mIoU for semantic segmentation on ADE20K, BLEU-1 score for vision-language (VL) reasoning on ScanQA, average score for robot learning on CortexBench, and Top-1 accuracy (\%) for classification on ImageNet-1K. Lower is better for depth estimation. Parameters of \cite{bolya2025perception} and \cite{maninis2025tips} are 1.9B and 1.1B, respectively.}
\label{tab:enhance_spatial_encoder}
\vspace{-0.08in}
\begin{adjustbox}{max width=\linewidth}
\begin{tabular}{lccccc}
\toprule
Method & Depth Estimation $\downarrow$ & Segmentation $\uparrow$ & VL Reasoning $\uparrow$ & Robot Learning $\uparrow$ & Classification $\uparrow$ \\
\midrule
TIPS~\cite{maninis2025tips} & 0.35 & 49.9 & 37.4 & 66.0 & 83.3 \\
\rowcolor{skyblue}
+\sname (ours) & \textbf{0.25} & \textbf{53.5} & \textbf{41.1} & \textbf{76.5} & \textbf{85.8} \\
\midrule
PE-Core~\cite{bolya2025perception} & 0.25 & 41.5 & 40.5 & 69.0 & 86.8 \\
\rowcolor{skyblue}
+\sname (ours) & \textbf{0.20} & \textbf{50.9} & \textbf{44.1} & \textbf{81.5} & \textbf{88.0} \\
PE-Spatial~\cite{bolya2025perception} & 0.26 & 49.3 & 40.2 & 68.7 & 85.7 \\
\rowcolor{skyblue}
+\sname (ours) & \textbf{0.19} & \textbf{56.3} & \textbf{43.6} & \textbf{81.2} & \textbf{87.5} \\
\bottomrule
\end{tabular}
\end{adjustbox}
\end{table*}

\vspace{0.05in}
\noindent{\bf Dataset Scalability.}
We analyze the impact of dataset sizes on depth estimation results from NYUd~\cite{silberman2012indoor} benchmark and semantic segmentation results from ADE20K~\cite{zhou2017scene} benchmark. 
As shown in \Cref{fig:ablation_datasize}, with matched training iterations (\ie, one epoch for 300K data), larger datasets yield consistent improvements, indicating robust scalability potential.

\vspace{0.05in}
\noindent{\bf Application to spatial-aware vision encoder.}
We investigate whether \sname further improves vision encoders that already possess strong spatial awareness. We evaluate the encoders on depth estimation, segmentation, vision-language reasoning, robot learning, and classification tasks. As shown in \Cref{tab:enhance_spatial_encoder}, \sname consistently improves performance across all benchmarks, demonstrating that our training approach is complementary to existing methods that enhance the spatial awareness of vision encoders.
\section{Conclusion}
\label{sec:conclusion}
In this paper, we have presented \sname, a framework to enhance the vision encoders by leveraging linguistic expressions of geometric and semantic information within images.
\sname uses LLM and dual-channel attention layers to exploit linguistic information into image representations, generates a multi-turn visual spatial reasoning dataset, and leverages them to improve the image representations.
Our experiments show that \sname consistently enhances the vision encoders on various downstream tasks that require a spatial understanding of images.
We hope that our work further facilitates future research on designing and enhancing vision encoders.

\clearpage  

%
%
\bibliographystyle{splncs04}
\bibliography{main}

\newpage
\appendix

\renewcommand{\thesection}{\Alph{section}}
\renewcommand{\thesubsection}{\Alph{section}.\arabic{subsection}}

\renewcommand{\theHsection}{appendix.\Alph{section}}
\renewcommand{\theHsubsection}{appendix.\Alph{section}.\arabic{subsection}}

\clearpage
\phantomsection
\section*{Appendix Contents}
\thispagestyle{plain}

\begingroup
\hypersetup{linkcolor=navyblue}
\small
\setlength{\parindent}{0pt}

\newcommand{\appsec}[2]{%
  \noindent\textbf{\hyperref[#2]{\ref*{#2}.~#1}}%
  \dotfill \hyperref[#2]{\pageref*{#2}} \par}
\newcommand{\appsubsec}[2]{%
  \noindent\hspace*{1.5em}\hyperref[#2]{\ref*{#2}~#1}%
  \dotfill \hyperref[#2]{\pageref*{#2}} \par}

\appsec{Implementation Details}{appendix:implementation_details}
\appsubsec{Training Details of Stage 1 \& 2}{appendix:stage1n2}
\appsubsec{Training Details of Stage 3}{appendix:stage3}
\appsubsec{Dense Prediction Tasks}{appendix:dense_prediction}
\appsubsec{3D Scene Understanding}{appendix:3d_understanding}
\appsubsec{Vision-based Robot Learning}{appendix:robot_learning}
\appsubsec{Image Classification Task}{appendix:image_classification}
\appsubsec{Image Retrieval Task}{appendix:image_retrieval}

\vspace{0.4em}

\appsec{Multi-view VQA Dataset}{appendix:multiview_vqa}

\vspace{0.4em}

\appsec{Multi-turn Visual Spatial Reasoning Dataset}{appendix:implementation_details_data}

\vspace{0.4em}

\appsec{Details of Ablation Study}{appendix:detail_ablation}

\vspace{0.4em}

\appsec{Additional Analysis}{appendix:analysis}
\appsubsec{Detailed Analysis on Reasoning Hierarchy}{appendix:analysis:reasoning}
\appsubsec{Detailed Analysis on Single-view and Multi-view Data}{appendix:analysis:single_and_multi_view}
\appsubsec{Detailed Analysis on Dual-channel Attention}{appendix:analysis:dual_channel}
\appsubsec{Detailed Results on Data Scalability}{appendix:analysis:data_scalability}
\appsubsec{Analysis on Bias Propagation in Reasoning Data}{appendix:analysis:bias}
\appsubsec{Additional Results on Multi-modal Large Language Models}{appendix:analysis:mllm}

\vspace{0.4em}

\appsec{Limitations}{appendix:limitations}

\endgroup
\clearpage

\section{Implementation Details}
\label{appendix:implementation_details}

\subsection{Training Details of Stage 1 \& 2}\label{appendix:stage1n2}
We train our multi-modal architecture with 4x NVIDIA Tesla A100s. In multi-modal architecture, we choose Qwen-2.0-7B~\cite{yang2024qwen2} as the LLM backbone and 2-layer MLP as the projector. In feature alignment pre-training (Stage 1), we train the projector on a BLIP-558K data~\cite{liu2024improved} for one epoch with a learning rate of 2e-3 and a batch size of 256. In visual instruction tuning (Stage 2), we fine-tune both the projector and the LLM backbone on the LLaVA-Instruct-158K dataset~\cite{liu2024improved} and our multi-view VQA dataset (described in \Cref{sec:method:alignment}) for one epoch with a batch size of 128.

\subsection{Training Details of Stage 3}\label{appendix:stage3}
In this stage, we adapt dual-channel attention layers in training vision encoders by introducing additional attention channels described in \Cref{sec:training_pipeline}. By applying dual-channel attention, the number of model parameters increased by 30\% in OpenCLIP and SigLIPv2 and by 25\% in DINOv2 and DINOv3, respectively.
We freeze the LLM decoder and fine-tune the vision encoder and projector on a multi-turn visual spatial reasoning dataset (described in \Cref{sec:method:dataset}) for one epoch with a learning rate of 2e-5 and a batch size of 128. We conduct hyperparameter search for the learning rate from 1e-6 to 1e-2.

\subsection{Dense Prediction Tasks}\label{appendix:dense_prediction}
From the vision encoder obtained through \sname, we performed depth estimation and semantic segmentation. We follow the same protocol as in DINOv2~\cite{oquab2023dinov2}, defining three primary hyperparameters for our linear probing setup: the learning rate, the number of output layers, and whether we concatenate the average-pooled patch token features with the class token. Concretely, we perform a grid search over learning rates in 1e-4 to 1e-1, choose the output layers from $\{1, 4\}$, and optionally concatenate average-pooled representations. We train each linear layer with SGD for 12500 iterations using random-resized-crop data augmentation. We then select the best hyperparameter combination on validation accuracy.

\subsection{3D Scene Understanding}\label{appendix:3d_understanding}
We evaluate whether \sname enables complex 3D-centric reasoning using the Lexicon3D~\cite{man2024lexicon3d} benchmark. Lexicon3D provides a unified probing framework that freezes visual backbones and attaches task-specific heads to evaluate vision-language reasoning, visual grounding, 3D semantic segmentation, and geometric correspondence. Following the Lexicon3D protocol, we extract features from 2D vision encoders and evaluate them on various 3D understanding tasks.

\vspace{0.05in}
\noindent{\bf Vision-Language Reasoning.}
To evaluate vision-language reasoning, we target the 3D visual question-answering (3D-VQA) on ScanQA~\cite{azuma2022scanqa} and SQA3D~\cite{ma2022sqa3d} datasets. We follow the 3D-LLM~\cite{hong20233dllm} architecture as our task head. Specifically, we use a Q-Former module~\cite{li2023blip} to project multi-view visual features into the input space of the language model. These projected features are then fed to the LLM (\eg, FlanT5~\cite{chung2024scaling}) for generating answers. We pre-train only the Q-Former projection module for 10 epochs using 3D-Language dataset~\cite{hong20233dllm} and fine-tune the module for 35 epochs using training split of ScanQA and SQA3D. We keep both the vision encoder and LLM frozen during training.

\vspace{0.05in}
\noindent{\bf Visual Grounding.}
To evaluate visual grounding with vision encoder, we target the object localization task based on text descriptions on the ScanRefer~\cite{chen2020scanrefer} dataset. We use an attention-based fusion head following Multi3DRefer~\cite{zhang2023multi3drefer}. The task head consists of a multi-layer attention module with 4 transformer layers that fuses visual and text embeddings. After projecting multi-view features to 3D space and extracting object features via average pooling within ground-truth bounding boxes, we apply cross-attention between object features and CLIP-encoded text descriptions. The fusion module outputs confidence scores for each object. We train the header for 30 epochs with cross-entropy loss.

\vspace{0.05in}
\noindent{\bf Geometric Understanding.}
To evaluate geometric understanding, we target the geometric correspondence task. We adopt a REGTR-style~\cite{yew2022regtr} transformer cross-encoder as the task head. The head process features from two partial point clouds to establish correspondences. After obtaining point correspondences through the transformer, we apply the Kabsch-Umeyama~\cite{kabsch1976solution, umeyama2002least} algorithm for closed-form estimation of rotation and translation parameters. We train the transformer head using partial scene registration benchmark~\cite{man2024lexicon3d} for 30 epochs using a combination of correspondence loss and transformation loss.

\vspace{0.05in}
\noindent{\bf 3D Semantic Understanding.}
To evaluate 3D semantic understanding, we target the point-wise semantic classification task on ScanNet~\cite{dai2017scannet}. We employ a linear probing head consisting of a single fully-connected layer followed by sigmoid activation:
$\textbf{y}=\text{Sigmoid}(\text{FC}(\textbf{x}))$, where $\textbf{x} \in \mathbb{R}^{N \times d}$ represents projected point features from multi-view images, $\textbf{y} \in \mathbb{R}^{N \times C}$ represents class probabilities for $C = 20$ semantic classes and $N$ is the number of points in each point cloud. The linear layer maps from feature dimension $d$ to the number of classes. We train the linear layer using ScanNet segmentation dataset with cross-entropy loss at learning rate 1e-4 for 20 epochs.

\subsection{Vision-based Robot Learning}\label{appendix:robot_learning}
We train the robot agents using 100 demos for each task. For training, we use keypoint augmentation~\cite{james2022q} for each demonstration, and use the end-effector controller with path planning as an action mode. We use the front camera of 224$\times$224 resolution without depth measurements. We evaluate the model 5 times by training with a pre-defined interval and report the mean of the best performance.

\subsection{Image Classification Task}\label{appendix:image_classification}
We train a linear classifier on top of the \texttt{[CLS]} token from the last feature of the vision encoder using the training split of ImageNet-1K~\cite{deng2009imagenet} dataset. Following the evaluation protocol of DINOv3~\cite{simeoni2025dinov3}, we employ SGD optimizer with momentum 0.9 and random-resized-crop data augmentation. We train the linear layer for 10 epochs with a batch size of 1024. We perform a grid search for the optimal learning rate, ranging from 1e-4 to 1e-1, selecting the best performing configuration.

\subsection{Image Retrieval Task}\label{appendix:image_retrieval}
We evaluate the image retrieval performance of vision encoders using a non-parametric retrieval approach. Specifically, we compute cosine similarity between the output \texttt{[CLS]} tokens of query and target images to establish ranking. For Oxford~\cite{radenovic2018revisiting}, Paris~\cite{radenovic2018revisiting}, and AmsterTime~\cite{yildiz2022amstertime} datasets, we resize images to $224 \times 224$ resolution, while for the Met~\cite{ypsilantis2021met} dataset, we resize to the nearest multiple of the patch size. All other setups follow evaluation protocols of each benchmark.

\newpage

\section{Multi-view VQA Dataset}\label{appendix:multiview_vqa}
We utilize multi-view data to inject rich 3D information into vision encoders. We find that proper instruction tuning is crucial for LLMs to stably transfer the 3D information to vision encoders. However, existing datasets are limited to enhance multi-view understanding, as most VQA datasets focus exclusively on single-view scenarios. We thereby construct a multi-view VQA dataset.

We consider both 3D datasets and ego-centric video data for our multi-view VQA construction. Specifically, we utilize ScanNet~\cite{dai2017scannet}, Mip-NeRF360~\cite{barron2022mip}, and MVImgNet~\cite{yu2023mvimgnet} for 3D data, and Ego4D~\cite{grauman2022ego4d} for ego-centric video data. From these datasets, we extract pairs of images that satisfy the following LPIPS~\cite{zhang2018unreasonable} constraint:
\begin{equation}
\label{eq:lpips}
0.35 \leq \text{LPIPS}(\rvx_{\mathtt{i}}, \rvx_{\mathtt{j}}) \leq 0.65, \text{where $\rvx_{\mathtt{i}}, \rvx_{\mathtt{j}} \in \{\rvx_{\mathtt{1}} \cdots \rvx_{\mathtt{N}}\}.$}
\end{equation}
This constraint effectively filters out outlier samples for meaningful multi-view learning. Given the selected image pairs, we utilize GPT-4o~\cite{achiam2023gpt} to generate three types of visual questions: (1) common VQA, (2) adversarial VQA, and (3) multi-choice VQA. These question types are designed to probe general knowledge understanding from multi-view visual inputs, thereby guiding the model to accurately process and answer multi-view visual questions. We provide specific prompts used for generating multi-view VQA data in \Cref{tab:template:multiview_vqa}.

\begin{table}[htbp]
    \centering\small
    \caption{Prompt examples for generating multi-view VQA data.}\label{tab:template:multiview_vqa}
    \vspace{-0.12in}
\resizebox{0.8\linewidth}{!}{
    \begin{tabular}{l}
    \toprule
\texttt{system$\_$prompt $=$[}\\
\quad ``\texttt{You are a helpful multimodal assistant.} \\
\quad \texttt{Generate question-answer pairs for given two images.} \\
\quad \texttt{Both images are came from same scene.} \\
\quad \texttt{When referring to the image, please call it the first image or the second image.}" \\
\texttt{]} \\
\texttt{general$\_$vqa$\_$prompt $=$[}\\
\quad ``\texttt{Please give me an exact question and answer by referring to the images.} \\
\quad \texttt{This is a common VQA.} \\
\quad \texttt{Create relevant question about these 2 images,} \\
\quad \texttt{referencing details that may only be visible if we consider both views.} \\
\quad \texttt{Then provide a concise, correct answer.} \\
\quad \texttt{The answer should be in length between 10 and 80 words.}" \\
\texttt{]} \\
\texttt{multi$\_$choice$\_$vqa$\_$prompt $=$[}\\
\quad ``\texttt{Please give me an exact question and answer by referring to the images.} \\
\quad \texttt{This is a multi-choice VQA.} \\
\quad \texttt{Create relevant question about these 2 images,} \\
\quad \texttt{referencing details that may only be visible if we consider both views.} \\
\quad \texttt{Then also generate 4 answer candidates,} \\
\quad \texttt{where only one candidate is correct and the others are very wrong.} \\
\quad \texttt{List candidates A to D or 1 to 4.} \\
\quad \texttt{The answer is the index of correct question.} \\
\quad \texttt{Each candidates should be in length between 5 and 20 words.} \\
\texttt{]} \\
    \bottomrule
    \end{tabular}}
\end{table}

\newpage

\section{Multi-turn Visual Spatial Reasoning Dataset}
\label{appendix:implementation_details_data}
We here provide a detailed implementation of the data generation pipeline and examples of multi-turn visual spatial reasoning.

We construct a multi-turn visual spatial reasoning dataset by associating each single-view image $\mathbf{x}$ or multi-view images $\{\mathbf{x}_{\texttt{1}} \cdots \mathbf{x}_{\texttt{N}}\}$ with 12 sequential QA turns. The first 5 turns focus on pixel-level view, prompting questions about point-wise depth or depth comparisons. The next 4 turns shift to object-level queries, referring to approximate bounding cubes (\ie, 3D bounding boxes) for each object. The next one turn addresses scene-level understanding, requiring holistic 3D interpretation. The last 2 turns are GPT-generated scene captions for given image input. 
For instance, the entire sequence of question-answer pairs for image $\mathbf{x}$ is described by
\begin{equation*}
\begin{aligned}
\text{Pixel-level} &: \bigl(Q_{\rvx}^{(1)}, A_{\rvx}^{(1)}\bigr)
\rightarrow
\cdots
\rightarrow
\bigl(Q_{\rvx}^{(5)}, A_{\rvx}^{(5)}\bigr) \rightarrow, \\
\quad
\text{Object-level} &: \bigl(Q_{\rvx}^{(6)}, A_{\rvx}^{(6)}\bigr)
\rightarrow
\cdots
\rightarrow
\bigl(Q_{\rvx}^{(9)}, A_{\rvx}^{(9)}\bigr) \rightarrow, \\
\quad
\text{Scene-level} &: \bigl(Q_{\rvx}^{(10)}, A_{\rvx}^{(10)}\bigr) \rightarrow, \\
\quad
\text{Scene Caption} &: \bigl(Q_{\rvx}^{(11)}, A_{\rvx}^{(11)}\bigr)
\rightarrow
\bigl(Q_{\rvx}^{(12)}, A_{\rvx}^{(12)}\bigr).
\end{aligned}
\end{equation*}
Each turn builds on the previous answers, allowing the LLM to engage in CoT reasoning. To extract 3D information for each image, we use the specialized vision models (\eg, depth and segmentation networks) and synthesize QA pairs that reflect the relevant 3D information, ensuring that the final scene-level query can integrate pixel-level and object-level details into a coherent spatial understanding.

\vspace{0.05in}
\noindent{\bf Filtering for Single-view Image.}
Generating visual spatial reasoning data requires multiple objects in an image. Therefore, selecting the appropriate images is necessary. Following SpatialVLM~\cite{chen2024spatialvlm} and SpatialRGPT~\cite{cheng2025spatialrgpt}, we adopt a CLIP-based open-vocabulary classification model~\cite{sun2023evaclip} to identify appropriate images with 100K samples from 314K samples of SA1B~\cite{kirillov2023segment}.
We provide the labels to get filtered images in \Cref{list:filtering}.
\begin{table}[htbp]
\centering
\caption{CLIP labels for filtering images.}
\vspace{-0.12in}
\label{list:filtering}
\resizebox{0.75\linewidth}{!}{
\begin{tabular}{cl}
\toprule
Label type & Labels \\
\midrule
\multirow{4}{*}{Positive labels} & ``\texttt{an iPhone photo of an indoor scene}" \\
 & ``\texttt{an iphone photo of an outdoor scene}"\\
 & ``\texttt{a DSLR photo of an indoor scene}"\\
 & ``\texttt{a DSLR of an outdoor scene}"\\
\midrule
\multirow{8}{*}{Negative labels} & ``\texttt{a close up shot of a single object}" \\
 & ``\texttt{a product displayed in front of a white background}" \\
 & ``\texttt{an artwork}" \\
 & ``\texttt{a painting}" \\
 & ``\texttt{a screenshot of a graphical user interface}" \\
 & ``\texttt{a piece of text}" \\
 & ``\texttt{a sketch}" \\
\bottomrule
\end{tabular}}
\vspace{-0.08in}
\end{table}

\vspace{0.05in}
\noindent{\bf Filtering for Multi-view Images.}
We apply LPIPS~\cite{zhang2018unreasonable} metric to 3D data (\eg, ScanNet~\cite{dai2017scannet} trainset) and ego-centric video data (\eg, Ego4D~\cite{grauman2022ego4d}) to obtain pairs of images that satisfy \Cref{eq:lpips}.
This constraint prevents sampling of image pairs that are either too dissimilar or overly redundant from the datasets.

\vspace{0.05in}
\noindent{\bf Point Cloud Processing.}
We process two types of input: (1) single-view and (2) multi-view. For a single-view image, we use the results of the segmentation and depth estimation to generate a 3D point cloud for objects in images. In particular, we use Depth-pro~\cite{bochkovskii2024depth} to perform metric depth estimation. For multi-view images, we obtain a 3D point cloud through VGGT~\cite{wang2025vggt}, which is a state-of-the-art 3D reconstruction model.
For each image input $\{\rvx_{\mathtt{1}} \cdots \rvx_{\mathtt{N}}\}$, we first select an image $\mathbf{x}_{\mathtt{i}}$, where $\rvx_{\mathtt{i}} \in \{\rvx_{\mathtt{1}} \cdots \rvx_{\mathtt{N}}\}$, among the image input and generate pixel-level data by randomly selecting the 2D coordinates of bounding boxes in $\rvx_{\mathtt{i}}$ and then extract the depth information. We also generate object and scene-level data by randomly selecting the bounding cubes obtained by using 3D point cloud.
We represent the bounding cubes in the canonical space, which is proposed by SpatialVLM~\cite{chen2024spatialvlm}.

\begin{table}[htbp]
    \centering\small
    \caption{Template examples for pixel-level VQA.}\label{tab:template:pixel}
    \vspace{-0.12in}
\resizebox{0.80\linewidth}{!}{
    \begin{tabular}{l}
    \toprule
\texttt{single$\_$point$\_$questions $=$[}\\
\quad ``\texttt{What is the depth value at pixel point [A]?}" \\
\quad ``\texttt{How far away is point [A]?}" \\
\quad ``\texttt{Tell me the depth of point [A].}" \\
\texttt{]} \\
\texttt{single$\_$point$\_$answers $= [$}\\
\quad ``\texttt{[X] away.}" \\
\quad ``\texttt{It is [X].}" \\
\quad ``\texttt{Depth value of point [A] is [X].}" \\
\texttt{]} \\
\texttt{close$\_$predicate$\_$questions $=$[}\\
\quad ``\texttt{Which point is close to a viewer? Point: [A], Point: [B].}" \\
\quad ``\texttt{Is point [A] closer than [B]?}" \\
\quad ``\texttt{Which point has a smaller depth value? Point [A] or Point [B]?}" \\
\quad ``\texttt{Compare the depth of point [A] and point [B].}" \\
\texttt{]} \\
\texttt{close$\_$true$\_$responses $=$[}\\
\quad ``\texttt{Yes, point [A] is closer to the viewer than point [B].}" \\
\quad ``\texttt{Indeed, point [A] has a smaller depth value than point [B].}" \\
\quad ``\texttt{Correct, point [A] is closer than point [B].}" \\
\texttt{]} \\
\texttt{close$\_$false$\_$responses $=$[}\\
\quad ``\texttt{No, point [A] is not closer than point [B].}" \\
\quad ``\texttt{In fact, point [B] is closer to the viewer than point [A].}" \\
\quad ``\texttt{Incorrect, point [B] has a smaller depth value than point [A].}" \\
\texttt{]} \\
    \bottomrule
    \end{tabular}}
\end{table}

\vspace{0.05in}
\noindent{\bf Pixel-level VQA Data.}
Pixel-level dataset has two types of QAs: (1) single point and (2) multi point QA. Single-point QA consists of questions that query depth values at specific coordinates on the image, and multi point QA involves comparing depth values between two different coordinates. To avoid generating excessively noisy data, all depth values are rounded to the third decimal place. We use a centimeter scale for depth values less than 0.5 meters while maintaining the template.
We provide examples of templates for each type of QA of this level in \Cref{tab:template:pixel}.

\begin{table}[htbp]
    \centering\small
    \caption{Template examples for object-level VQA.}\label{tab:template:object}
    \vspace{-0.12in}
\resizebox{0.8\linewidth}{!}{
    \begin{tabular}{l}
    \toprule
\texttt{bounding$\_$cube$\_$questions $=$[}\\
\quad ``\texttt{Identify [A] and [B]}" \\
\quad ``\texttt{What is the center of the 3d bounding box coordinate for [A]?}" \\
\texttt{]} \\
\texttt{bounding$\_$cube$\_$answers $=$[}\\
\quad ``\texttt{[X]}" \\
\quad ``\texttt{Center: [X]}" \\
\quad ``\texttt{[A] in [X] and [B] in [Y]}" \\
\texttt{]} \\
\texttt{left$\_$predicate$\_$questions $=$[}\\
\quad ``\texttt{Is the [A] to the left of the [B] from the viewer's perspective?}" \\
\quad ``\texttt{Does the [A] appear on the left side of the [B]?}" \\
\quad ``\texttt{Can you confirm if the [A] is positioned to the left of the [B]?}" \\
\texttt{]} \\
\texttt{left$\_$true$\_$responses $=$[}\\
\quad ``\texttt{Yes, the [A] is to the left of the [B].}" \\
\quad ``\texttt{Indeed, the [A] is positioned on the left side of the [B].}" \\
\quad ``\texttt{Correct, you'll find the [A] to the left of the [B].}" \\
\texttt{]} \\
\texttt{left$\_$false$\_$responses $=$[}\\
\quad ``\texttt{No, the [A] is not to the left of the [B].}" \\
\quad ``\texttt{In fact, the [A] is either to the right of or directly aligned with the [B].}" \\
\quad ``\texttt{Incorrect, the [A] is not on the left side of the [B].}" \\
\texttt{]} \\
    \bottomrule
    \end{tabular}}
\end{table}

\begin{table}[htbp]
    \centering\small
    \caption{Template examples for scene-level VQA.}\label{tab:template:scene}
    \vspace{-0.12in}
\resizebox{0.8\linewidth}{!}{
    \begin{tabular}{l}
    \toprule
\texttt{distance$\_$questions $=$[}\\
\quad ``\texttt{What is the distance between the [A] and the [B]?}" \\
\quad ``\texttt{How far is the [A] from the [B]?}" \\
\quad ``\texttt{How distant is the [A] from the [B]?}" \\
\quad ``\texttt{Measure the distance from the [A] to the [B].}" \\
\texttt{]} \\
\texttt{distance$\_$answers $=$[}\\
\quad ``\texttt{[X]}" \\
\quad ``\texttt{the [A] and the [B] are [X] apart.}" \\
\quad ``\texttt{They are [X] apart.}" \\
\quad ``\texttt{The distance of the [A] from the [B] is [X].}" \\
\texttt{]} \\
    \bottomrule
    \end{tabular}}
\end{table}

\vspace{0.05in}
\noindent{\bf Object-level VQA Data.}
Object-level dataset has two types of QAs: (1) predicting a bounding cube of an object from the bounding box of the object, and (2) predicting the relative positional relationship between two objects. We provide examples of templates for each type of QA of this level in \Cref{tab:template:object}.

\vspace{0.05in}
\noindent{\bf Scene-level VQA Data.}
Scene-level dataset has single type of QA: predicting the 3D relative distance between two objects. We provide examples of templates for each type of QA of this level in \Cref{tab:template:scene}.

\vspace{0.05in}
\noindent{\bf Expand Viewpoints in Multi-view Data.}
Through the aforementioned process, we obtain multi-view reasoning data for 2-view images. We denote these obtained views as anchor views. To extend beyond 2-view configurations, we additionally sample interpolated frames between the anchor views and validate whether the VQA pairs generated for the anchor views remain valid for these new viewpoints using GPT-4o. Specifically, if the existing VQA pairs are verified as correct for more than half of the interpolated views, we incorporate these interpolated views as additional viewpoints. This approach enables us to extend the 2-view input to arbitrary multi-view configurations. Among our 200K multi-view samples, we have 160K 2-view samples, 30K 4-view samples, and 10K 8-view samples.

\section{Details of Ablation Study}\label{appendix:detail_ablation}
We here provide a detailed implementation of the ablation study in \Cref{tab:analysis_curriculum}.

Our key hypothesis is that language supervision, particularly through LLM-based supervised fine-tuning, can effectively distill rich 3D information into vision encoders. To validate this, we investigate whether LLM provides superior supervision compared to pixel-level alternatives. We align various headers with vision encoders following the \sname framework, then fine-tune the vision encoder with dual-channel attention. We evaluate each enhanced vision encoder on ImageNet-1K~\cite{deng2009imagenet} image classification, ADE20K~\cite{zhou2017scene} semantic segmentation, and NYUd~\cite{silberman2012indoor} monocular depth estimation. As shown in \Cref{tab:analysis_curriculum}, pixel-level supervision leads to catastrophic forgetting, while language supervision preserves pre-trained knowledge. This validates our hypothesis that language serves as an effective modality for transferring dense and hierarchical spatial information.

For all experiments, we fine-tune the vision encoder with fixed 300K samples extracted from our multi-turn visual reasoning dataset, except for the VGGT experiment. We choose DINOv2-ViT-L/14 as a vision encoder architecture, with following evaluation protocols for each downstream task detailed in \Cref{appendix:implementation_details}. The specific implementation for each header-based fine-tuning approach is provided in following paragraphs:

\vspace{0.05in}
\noindent{\bf SAM Decoder.}
We adopt the SAM decoder as a header and introduce an MLP layer to match dimensions with the vision encoder. Following the \sname training strategy, we first align only the MLP layer using 300K samples from SA1B~\cite{kirillov2023segment} dataset. Subsequently, we apply dual-channel attention to the vision encoder and fine-tune it using 300K segmentation samples from our multi-turn visual reasoning dataset, which is also sampled from SA1B dataset.

\vspace{0.05in}
\noindent{\bf VGGT Decoder.}
VGGT~\cite{wang2025vggt} is a state-of-the-art 3D reconstruction model that employs DINOv2-ViT-L/14-reg~\cite{darcet2023register} as a feature extractor. Building upon this off-the-shelf pipeline, we apply dual-channel attention to the vision encoder and perform fine-tuning. We utilize 300K 3D data samples from Co3D~\cite{reizenstein2021co3d} for training.

\vspace{0.05in}
\noindent{\bf Linear Layers.}
We consider two different pixel-level modalities as input for linear layers: (1) depth and (2) segmentation. As linear layers are randomly initialized, we first train the linear layer while freezing the vision encoder. We use 300K samples from SA1B to train the linear layer, then apply dual-channel attention to the vision encoder and fine-tune the vision encoder with 300K samples from our reasoning data. For depth data, we use depth maps obtained through Depth-Pro~\cite{bochkovskii2024depth} on SA1B and a subset of our reasoning dataset. For segmentation data, we follow the same data configuration in SAM decoder experiment.

\vspace{0.05in}
\noindent{\bf LLM (Ours).}
We use Qwen-2.0-7B~\cite{yang2024qwen2} as the LLM backbone. We train the projector with 300K SA1B and fine-tune the vision encoder and projector with 300K samples from our reasoning data. We follow all other training setup described in \Cref{appendix:implementation_details}.

\section{Additional Analysis}\label{appendix:analysis}

\subsection{Detailed Analysis on Reasoning Hierarchy}\label{appendix:analysis:reasoning}
In this section, we investigate which components of the multi-turn visual reasoning data contribute most significantly to the performance of \sname. We provide a detailed analysis.

\begin{table}[htbp]
    \centering
    \scriptsize
    \setlength{\tabcolsep}{3pt}
    \renewcommand{\arraystretch}{1.0}
    \caption{Effect of reasoning hierarchy.}
    \resizebox{\linewidth}{!}{
    \begin{tabular}{lcccccccccc}
        \toprule
        & \multicolumn{4}{c}{Depth $\downarrow$} & \multicolumn{4}{c}{Segmentation $\uparrow$} & \multicolumn{2}{c}{Classification $\uparrow$} \\
        \cmidrule(lr){2-5} \cmidrule(lr){6-9} \cmidrule(lr){10-11}
        Method & \multicolumn{2}{c}{OpenCLIP} & \multicolumn{2}{c}{DINOv2} 
               & \multicolumn{2}{c}{OpenCLIP} & \multicolumn{2}{c}{DINOv2} & OpenCLIP & DINOv2 \\
        & lin. & DPT & lin. & DPT & lin. & +ms & lin. & +ms & & \\
        \midrule
        Pre-trained & 0.56 & 0.41 & 0.38 & 0.29 & 39.1 & 45.7 & 47.7 & 53.1 & 83.9 & 86.3 \\
        \midrule
        Pix           & 0.52 & 0.40 & 0.34 & 0.29 & 39.6 & 46.3 & 48.2 & 53.4 & 84.0 & 86.6 \\
        Obj           & 0.53 & 0.41 & 0.37 & 0.30 & 39.4 & 46.3 & 48.0 & 53.3 & 84.7 & 87.2 \\
        Scene         & 0.53 & 0.41 & 0.38 & 0.32 & 39.2 & 45.9 & 47.7 & 53.3 & 84.5 & 87.1 \\
        \midrule
        Pix + Obj     & 0.44 & 0.39 & 0.35 & 0.28 & 39.8 & 46.6 & 48.8 & 53.5 & 84.7 & 87.3 \\
        Pix + Scene   & 0.46 & 0.40 & 0.36 & 0.28 & 39.5 & 46.5 & 48.5 & 53.4 & 84.4 & 87.2 \\
        Obj + Scene   & 0.51 & 0.42 & 0.39 & 0.31 & 39.5 & 46.5 & 47.6 & 53.3 & 85.0 & 87.4 \\
        \midrule
        \rowcolor{skyblue}
        Pix + Obj + Scene
                      & \textbf{0.42} & \textbf{0.39} & \textbf{0.32} & \textbf{0.27} 
                      & \textbf{40.0} & \textbf{46.9} & \textbf{49.2} & \textbf{54.2} & \textbf{85.1} & \textbf{87.6} \\
        \bottomrule
    \end{tabular}
    }
    \label{tab:detailed_hierarchy}
\end{table}

\vspace{0.05in}
\noindent{\bf Setup.}
We explore which levels of the reasoning hierarchy have an impact on the performance of \sname by measuring the performance of vision encoders fine-tuned with different combinations of reasoning levels. For all experiments, we fix the sample size at 100K and ensure identical ratio for each combination. We evaluate monocular depth estimation on NYUd (RMSE), semantic segmentation on ADE20K (mIoU), and classification on ImageNet-1K (Top-1 accuracy). We use ViT-L/14 as a vision encoder architecture in all experiments. All other setups are the same as described in \Cref{appendix:implementation_details}.

\vspace{0.05in}
\noindent{\bf Results.}
As shown in \Cref{tab:detailed_hierarchy}, we observe that pixel-level QA and its combinations remark superior performance in dense prediction tasks, indicating pixel-level QA aids in higher-level understanding. We also observe that object-level QA and its combinations achieve strong improvements in classification. The results highlights that the combination with all levels achieves the best performance across all tasks, validating the effectiveness of our hierarchical reasoning.

\subsection{Detailed Analysis on Single-view and Multi-view Data}\label{appendix:analysis:single_and_multi_view}

In this section, we investigate the effect of single-view and multi-view data on various downstream tasks. We provide a detailed analysis.

\begin{table*}[h]
\centering
\small
\captionof{table}{
Effect of single-view and multi-view data across diverse tasks.}
\label{tab:single_multi_view_effect}
\vspace{-0.08in}
\begin{adjustbox}{max width=1.0\linewidth}
\begin{tabular}{lcccccccccccc}
\toprule
\multirow{2}{*}{Model} & \multirow{2}{*}{SV} & \multirow{2}{*}{MV} & \multirow{2}{*}{Cls $\uparrow$} & \multirow{2}{*}{Seg $\uparrow$} & \multirow{2}{*}{Depth $\downarrow$} &
\multicolumn{2}{c}{VLR} & VG & \multicolumn{2}{c}{GU} & \multicolumn{2}{c}{3D SU} \\
\cmidrule(lr){7-8} \cmidrule(lr){9-9} \cmidrule(lr){10-11} \cmidrule(lr){12-13}
 & & & & & & ScanQA $\uparrow$ & SQA3D $\uparrow$ & ScanRef $\uparrow$ & RR@0.05m $\uparrow$ &
RTE $\downarrow$ & Acc $\uparrow$ & mIoU $\uparrow$ \\
\midrule

SigLIPv2 & - & - & 89.1 & 42.8 & 0.51 & 38.1 & 48.5 & 51.4 & 47.8 & 0.28 & 47.7 & 9.2 \\

& +200K & +100K & \textbf{90.2} & 44.7 & 0.41 & 40.5 & 50.0 & 56.6 & 84.1 & 0.18 & 77.7 & 51.8 \\

& +150K & +150K & 90.0 & 44.9 & 0.39 & 40.6 & 50.1 & 56.6 & 84.9 & 0.16 & 80.2 & 52.4 \\

& +100K & +200K & 90.0 & \textbf{45.1} & \textbf{0.39} & \textbf{40.8} & \textbf{50.1} & \textbf{56.8} &
\textbf{86.4} & \textbf{0.15} & \textbf{81.0} & \textbf{55.5} \\
\midrule

DINOv3 & - & - & 88.4 & 55.9 & 0.31 & 40.6 & 51.4 & 56.2 & 86.9 & 0.10 & 91.1 & 69.1 \\

& +200K & +100K & 90.2 & 59.5 & 0.27 & 43.1 & 54.7 & 61.1 & 96.0 & 0.08 & 91.4 & 69.7 \\

& +150K & +150K & \textbf{90.3} & 59.6 & 0.26 & 43.1 & \textbf{55.0} & 61.1 & 96.9 & 0.07 & 91.6 & 70.2 \\

& +100K & +200K & 90.2 & \textbf{59.7} & \textbf{0.25} & \textbf{43.3} & 54.9 & \textbf{61.1} &
\textbf{97.5} & \textbf{0.06} & \textbf{91.9} & \textbf{70.6} \\
\bottomrule
\end{tabular}
\end{adjustbox}
\vspace{-0.4cm}
\end{table*}

\vspace{0.05in}
\noindent{\bf Setup.}
We explore the effect of single-view and multi-view data by fine-tuning the vision encoder with different proportions of our reasoning data. With fixed total samples, \ie, 300K from multi-turn spatial reasoning data, we train the vision encoders and evaluate them on classification (Cls) on ImageNet-1K, segmentation on ADE20K, depth estimation on NYUd, and 3D-centric tasks on Lexicon3D~\cite{man2024lexicon3d}. We use ViT-g/16 and ViT-7B/16 as the architecture of SigLIPv2 and DINOv3, respectively. All other setups are the same as described in \Cref{appendix:implementation_details}.

\vspace{0.05in}
\noindent{\bf Results.}
As shown in \Cref{tab:single_multi_view_effect}, we observe that multi-view reasoning data leads to improvements in tasks which require spatial knowledge such as depth estimation, segmentation, geometric understanding (GU), and 3D semantic understanding (3D SU). Following the size of multi-view data, SigLIPv2's GU registration recall improves from 84.1\% to 86.4\%, and 3D SU mIoU improves from 51.8\% to 55.5\%. These results demonstrate that multi-view reasoning data can effectively enhance 3D understanding of the vision encoder.

\subsection{Detailed Analysis on Dual-channel Attention}\label{appendix:analysis:dual_channel}

\begin{figure*}[h]
  \centering
    \centerline{\includegraphics[width=0.45\textwidth]{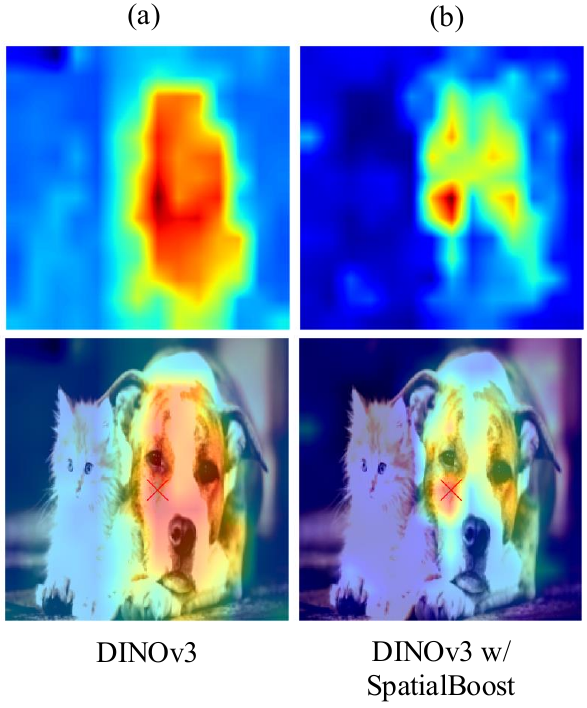}}
    \caption{\textbf{Qualitative results for dual-channel attention.} We visualize attention heatmap from (a) DINOv3 and (b) \sname DINOv3. We rollout attention layers for cosine similarity between patches. Red cross denotes a query patch. We visualize pure attention heatmap (top) and RGB overlayed version (bottom).}
   \label{fig:dual_channel_figure}
\end{figure*}

We provide quantitative and qualitative results for dual-channel attention (see \Cref{tab:quantitative_dual_channel} and \Cref{fig:dual_channel_figure}).

\begin{table}[htbp]
\centering
\small
\caption{Quantitative results of dual-channel attention.}
\begin{tabular}{lccc}
\toprule
Method & Classification $\uparrow$ & Segmentation $\uparrow$ & Depth estimation $\downarrow$ \\
\midrule
DINOv2 (Pre-trained)          & 86.3 & 47.7 & 0.38 \\
\midrule
Full Fine-tuning    & 79.5 & 49.4 & 0.31 \\
LoRA                & 81.7 & 49.0 & 0.32 \\
Dual-Channel Attn.  & 87.6 & 49.2 & 0.32 \\
\bottomrule
\end{tabular}
\label{tab:quantitative_dual_channel}
\end{table}

\vspace{0.05in}
\noindent{\bf Setup.}
We evaluate different fine-tuning methodologies while fixing the reasoning data sample size at 100K. Performance is measured on ImageNet-1K classification (accuracy), ADE20K segmentation (mIoU), and NYUd depth estimation (RMSE). All experiments utilize DINOv2 with ViT-L/14 architecture.

\vspace{0.05in}
\noindent{\bf Results.} As shown in \Cref{tab:quantitative_dual_channel}, we find that full fine-tuning and LoRA similarly exhibits performance drops in classification. In contrast, dual-channel attention shows consistent performance improvements across all tasks. This indicates that dual-channel attention effectively enhances spatial capabilities while preventing overfitting to spatial-specific features, maintaining the generalization ability. Partial results of \Cref{tab:quantitative_dual_channel} are visualized in \Cref{fig:ablation_class}.

\subsection{Detailed Results on Data Scalability}\label{appendix:analysis:data_scalability}

We provide more detailed results for data scalability. In \Cref{tab:detail_data_scalability}, \sname improves SigLIPv2 and DINOv3 in all tasks.

\begin{table}[htbp]
    \centering
    \small
    \caption{Data scalability on classification, segmentation, and depth estimation.}
    \begin{tabular}{llccc}
    \toprule
    Model & Sample size & Classification $\uparrow$ & Segmentation $\uparrow$ & Depth estimation $\downarrow$ \\
    \midrule
    SigLIPv2 & Pre-trained & 89.1 & 42.8 & 0.51 \\
    \midrule
    \multirow[t]{3}{*}{+\sname}
    & 50K  & 89.5 & 43.2 & 0.44 \\
    & 100K & 89.7 & 44.5 & 0.42 \\
    & \cellcolor{skyblue}{300K} & \cellcolor{skyblue}{90.0} & \cellcolor{skyblue}{45.1} & \cellcolor{skyblue}{0.39} \\
    \midrule
    DINOv3 & Pre-trained & 88.4 & 55.9 & 0.31 \\
    \midrule
    \multirow[t]{3}{*}{+\sname}
    & 50K  & 88.6 & 56.8 & 0.29 \\
    & 100K & 90.0 & 58.3 & 0.28 \\
    & \cellcolor{skyblue}{300K} & \cellcolor{skyblue}{90.2} & \cellcolor{skyblue}{59.7} & \cellcolor{skyblue}{0.25} \\
    \bottomrule
    \end{tabular}
    \label{tab:detail_data_scalability}
\end{table}

\subsection{Analysis on Bias Propagation in Reasoning Data}\label{appendix:analysis:bias}

We provide an analysis of bias in vision foundation models used to generate spatial reasoning data.

\begin{table*}[h]
\centering
\small
\captionof{table}{Comparison between VFM-based and GT-based reasoning data.}
\label{tab:vfm_vs_gt}
\vspace{-0.08in}
\begin{adjustbox}{max width=0.85\linewidth}
\begin{tabular}{lcccc}
\toprule
Method & Cls $\uparrow$ & Seg $\uparrow$ & Depth $\downarrow$ & VLR $\uparrow$ \\
\midrule
DINOv2 & 86.3 & 47.7 & 0.38 & 39.2 \\
+VFM-based & 87.5 & 48.7 & 0.34 & 39.6 \\
+GT-based  & 87.5 & 48.8 & 0.34 & 36.9 \\
\midrule
$\Delta$ (VFM $-$ GT) & \textbf{0.0} & \textbf{-0.1} & \textbf{0.0} & \textbf{0.0} \\
\bottomrule
\end{tabular}
\end{adjustbox}
\vspace{-0.4cm}
\end{table*}

\vspace{0.05in}
\noindent{\bf Setup.}
We explore the effect of bias propagation from vision foundation models (\eg, SAM, Depth-pro) used to generate spatial reasoning data. With fixed 100K ScanNet~\cite{dai2017scannet} single-view samples, we generate reasoning data based on 3D metadata extracted from vision foundation models (VFM-based) and ScanNet ground-truth annotation (GT-based). We then fine-tune the vision encoder and evaluate the performance on ImageNet-1K classification (Cls), ADE20K segmentation, NYUd depth estimation, and ScanQA vision-language reasoning (VLR).

\vspace{0.05in}
\noindent{\bf Results.}
As shown in \Cref{tab:vfm_vs_gt}, we observe that the performance between VFM-based and GT-based is negligible. The results demonstrate that the effect of bias propagation is marginal in our reasoning data pipeline.

\subsection{Additional Results on Multi-modal Large Language Models}\label{appendix:analysis:mllm}

We provide results of application our framework on Multi-modal Large Language Models (MLLM).

\begin{table*}[h]
\centering
\small
\captionof{table}{Effect of \sname on MLLM visual encoders.}
\label{tab:mllm_encoder}
\vspace{-0.08in}
\begin{adjustbox}{max width=0.75\linewidth}
\begin{tabular}{lcccc}
\toprule
Method & \#Params & Cls $\uparrow$ & Seg $\uparrow$ & Depth $\downarrow$ \\
\midrule
InternViT-6B-v2.5 & 5.5B & 86.6 & 39.4 & 0.46 \\
\rowcolor{skyblue}
+\sname (Ours) & 6.0B & \textbf{89.1} & \textbf{48.5} & \textbf{0.35} \\
\midrule
Qwen3-VL-VE & 0.6B & 87.9 & 40.8 & 0.44 \\
\rowcolor{skyblue}
+\sname (Ours) & 0.7B & \textbf{89.3} & \textbf{44.3} & \textbf{0.36} \\
\bottomrule
\end{tabular}
\end{adjustbox}
\vspace{-1cm}
\end{table*}

\begin{table*}[h]
\centering
\small
\captionof{table}{Effect of \sname on MLLM VQA performance.}
\label{tab:mllm_vqa}
\vspace{-0.08in}
\begin{adjustbox}{max width=\linewidth}
\begin{tabular}{lccccccc}
\toprule
Method & MMMU & RealWorldQA & OCRBench & DocVQA & BLINK & MUIRBench & ERQA \\
\midrule
InternVL3-38B & 70.1 & 75.6 & 886 & 95.4 & 64.0 & 63.8 & 42.8 \\
\rowcolor{skyblue}
+\sname (Ours) & \textbf{70.8} & \textbf{75.9} & \textbf{894} & 95.4 & \textbf{69.2} & \textbf{70.7} & \textbf{49.3} \\
\midrule
Qwen3-VL-32B-Instruct & 76.0 & 79.0 & 895 & 96.9 & 67.3 & 72.8 & 48.8 \\
\rowcolor{skyblue}
+\sname (Ours) & \textbf{76.4} & \textbf{79.6} & \textbf{909} & \textbf{97.1} & \textbf{70.8} & \textbf{76.4} & \textbf{51.5} \\
\bottomrule
\end{tabular}
\end{adjustbox}
\end{table*}

\vspace{0.05in}
\noindent{\bf Setup.}
We apply \sname on the vision encoders of InternVL3~\cite{zhu2025internvl3} and Qwen3-VL~\cite{bai2025qwen3vl}. With fixed 300K samples from our reasoning data, we fine-tune the vision encoder and evaluate linear probing for ImageNet-1K classification, ADE20K segmentation, and NYUd depth estimation. Additionally, we evaluate the performance of MLLM with \sname encoder on VQA tasks targeting multi-modal reasoning~\cite{yue2024mmmu}, real world comprehension~\cite{realworldqa}, OCR and document understanding~\cite{liu2024ocrbench,mathew2021docvqa}, multi-image comprehension~\cite{fu2024blink,wang2024muirbench}, and embodied reasoning~\cite{team2025geminirobotics}.

\vspace{0.05in}
\noindent{\bf Results.}
As shown in \Cref{tab:mllm_encoder}, we observe that \sname produces notable performance gain in the vision encoders of Qwen3-VL and InternVL3. For example, InternViT-6B-v2.5 with \sname raises the mIoU 39.4 to 48.5 on segmentation task. In \Cref{tab:mllm_vqa}, we observe that \sname yield consistent performance improvements on diverse VQA tasks. For instance, Qwen3-VL with \sname vision encoder rises the score from 72.8 to 76.4 on MUIRBench and  from 48.8 to 51.5 on ERQA.

\clearpage

\section{Limitations}\label{appendix:limitations}
Although \sname demonstrates strong performance across a wide range of benchmarks, our proposed pipeline for constructing the visual spatial reasoning dataset relies on vision models, which is a limitation also shared by many recent works on spatial VQA. Leveraging the remarkable capabilities of recent vision foundation models, we empirically show through controlled experiments that errors from these models do not meaningfully propagate into the fine-tuned vision encoder. Nevertheless, fundamentally addressing this concern would require large-scale ground-truth spatial annotations. However, collecting and annotating such data remains highly costly and largely unexplored. Therefore, we leave the curation of large-scale spatial ground-truth data and the development of more accurate vision foundation models as promising directions for extending our pipeline, and encourage the community to pursue these efforts.

\end{document}